\documentclass{article}

\PassOptionsToPackage{numbers, compress}{natbib}
\usepackage[preprint]{neurips_2026}

\usepackage[utf8]{inputenc} 
\usepackage[T1]{fontenc}    
\usepackage{hyperref}       
\usepackage{url}            
\usepackage{booktabs}       
\usepackage{amsfonts}       
\usepackage{nicefrac}       
\usepackage{microtype}      
\usepackage{xcolor}         

\usepackage{multirow}
\usepackage{amsmath}
\usepackage{amssymb}
\usepackage{graphicx}
\usepackage{colortbl}
\usepackage{wrapfig}
\usepackage{algorithm}
\usepackage{algpseudocodex}

\definecolor{myframepink}{RGB}{72,138,176}

\hypersetup{
    colorlinks=true,       
    allcolors=myframepink  
}

\title{System-Anchored Knee Estimation for Low-Cost Context Window Selection in PDE Forecasting}

\author{
Wenshuo Wang$^{1}$, Fan Zhang$^{2,3}$\thanks{Corresponding author.} \\
$^{1}$ School of Future Technology, South China University of Technology, China \\
$^{2}$ State Key Laboratory of Ocean Sensing \& Ocean College, Zhejiang University, China \\
$^{3}$ Kavli Institute for Astrophysics and Space Research, Massachusetts Institute of Technology, USA \\
\texttt{202364870251@mail.scut.edu.cn}, \texttt{f.zhang@zju.edu.cn}
}

\begin{document}

\maketitle

\begin{abstract}
Autoregressive neural PDE simulators predict the evolution of physical fields one step at a time from a finite history, but low-cost context-window selection for such simulators remains an unformalized problem. Existing approaches to context-window selection in time-series forecasting include exhaustive validation, direct low-cost search, and system-theoretic memory estimation, but they are either expensive, brittle, or not directly aligned with downstream rollout performance. We formalize explicit context-window selection for fixed-window autoregressive neural PDE simulators as an independent low-cost algorithmic problem, and propose \textbf{System-Anchored Knee Estimation (SAKE)}, a two-stage method that first identifies a small structured candidate set from physically interpretable system anchors and then performs knee-aware downstream selection within it. Across all eight PDEBench families evaluated under the shared \(L\in\{1,\dots,16\}\) protocol, SAKE is the strongest overall matched-budget low-cost selector among the evaluated methods, achieving 67.8\% Exact, 91.7\% Within-1, 6.1\% mean regret@knee, and a cost ratio of 0.051 (94.9\% normalized search-cost savings).
\end{abstract}

\section{Introduction}

Autoregressive neural simulators for PDEs predict the evolution of physical fields by advancing one time step at a time from a finite history of past states~\cite{mccabe2024mpp,hao2024dpot}. In this setting, choosing the context window length is a fundamental yet surprisingly under-studied problem: the window determines how much temporal information the simulator receives, but also how much training, evaluation, and tuning cost must be paid. Existing work typically fixes this window as a task-specific hyperparameter or a priori~\cite{li2020fno,mccabe2024mpp}, or folds it implicitly into memory-based architectures~\cite{ruiz2024memory}. Other work examines its effect only through ablations, rather than treating context-window selection itself as an independent optimization problem~\cite{tran2021ffno,lippe2023pderefiner}. As a result, a basic question remains open: can we identify the performance-critical window for a given autoregressive neural simulator without paying the cost of exhaustive search? The problem is challenging because the best window depends jointly on the underlying system dynamics, the simulator architecture, and the downstream rollout objective.

More broadly, context-window selection has been studied in time-series forecasting, where existing approaches can be roughly grouped into three families~\cite{leites2024lag}. The first is exhaustive validation over candidate windows~\cite{leites2024lag}. The second is direct low-cost search, which replaces full evaluation with cheap pilots, early stopping, or multi-fidelity resource allocation over a generic shortlist of candidate windows~\cite{li2018hyperband,li2020asha,snoek2012practical,kandasamy2017mfbo,falkner2018bohb}. The third is system-theoretic memory estimation, which infers a characteristic memory or embedding scale from the underlying dynamics and uses it to choose or bracket plausible windows~\cite{takens2006strange,fraser1986independent,kennel1992falseneighbors,cao1997practical}. Yet none of them fully addresses our setting. Exhaustive validation is accurate but expensive~\cite{leites2024lag}; direct search is cheaper but often system-agnostic and brittle~\cite{li2018hyperband,li2020asha}; and memory estimates are physically meaningful but do not directly determine the window that optimizes downstream rollout performance in neural PDE simulators~\cite{takens2006strange,fraser1986independent,cao1997practical}.

We therefore formalize explicit context-window selection for autoregressive neural PDE simulators as an independent low-cost algorithmic problem, and propose \textbf{System-Anchored Knee Estimation (SAKE)}. SAKE is a two-stage method. It first extracts physically interpretable system anchors from clean-system dynamics to identify a small but structured region of plausible windows. It then performs knee-aware downstream selection only within this anchor-induced candidate set, using a small number of task-level evaluations to choose the final window. By coupling system priors with lightweight downstream selection, SAKE avoids both the cost of exhaustive sweeps and the brittleness of generic low-cost search, providing a more reliable and efficient way to determine performance-critical context windows.

We evaluate SAKE on two representative PDEBench tasks, four autoregressive simulator backbones, and three training seeds in the main body, and complement these case-level results with an appendix-wide aggregate over all eight PDEBench families under a shared \(L\in\{1,\dots,16\}\) protocol. Across this fixed-window setting, we measure exact oracle-knee recovery, absolute window error, regret at the knee, and search cost. The aggregate results show that SAKE has the strongest overall matched-budget profile among the compared selectors in this scope: it achieves 67.8\% Exact, 91.7\% Within-1, 6.1\% mean regret@knee, and a 0.051 cost ratio, outperforming the generic shortlist and ASHA baselines without using a larger search budget. These gains do not come from more aggressive search. Instead, they come from coupling system-informed anchors with lightweight downstream selection.
In summary, the contributions of this paper are:
\begin{itemize}
    \item We are the first to formalize explicit context-window selection for autoregressive neural PDE simulators as an independent low-cost algorithmic problem. This problem is important because exhaustive context-window tuning can be computationally expensive in practice.
    \item We propose SAKE, a two-stage method that combines physically interpretable system anchors with knee-aware downstream selection.
    \item We empirically validate the effectiveness of SAKE across multiple PDE datasets, simulator backbones, and training seeds. In the fixed-window autoregressive setting studied here, SAKE provides a more reliable and efficient low-cost selector of context windows than the generic shortlist baselines considered in the main text.
\end{itemize}

\section{Related Work}

\subsection{Temporal dependence and historical information in neural simulators}

Recent work on time-dependent neural PDE solvers increasingly treats temporal dependence as a central challenge rather than a secondary implementation detail. Standard neural operators have been found to struggle with temporal dynamics beyond the horizons seen during training, architecture papers now target time-evolving dynamics directly, long-rollout studies make stable temporal prediction a primary concern, and recent benchmarks explicitly emphasize rollout metrics for temporal generalization \cite{diab2025temporal,zhang2024sinenet,lippe2023pderefiner,koehler2024apebench}. This importance of historical information is made especially explicit by \emph{On the Benefits of Memory for Modeling Time-Dependent PDEs}, which points out that many existing approaches are effectively Markovian and shows that using past states can substantially improve performance when the system is partially observed, noisy, or rich in high-frequency content \cite{ruiz2024memory}. Yet, despite this growing recognition, historical information is still usually handled implicitly rather than selected explicitly: many methods condition on only the current state or immediate past state, while others absorb memory into a dedicated architecture instead of treating the amount of accessible history as a separate modeling decision \cite{ruiz2024memory,diab2025temporal}. When the role of history is examined more directly, it is typically through architecture ablations, rollout-strategy studies, or benchmark analyses aimed at temporal extrapolation and stability, rather than through a procedure that explicitly selects a context window for a fixed autoregressive neural simulator \cite{diab2025temporal,lippe2023pderefiner,koehler2024apebench}.

The gap, therefore, is not that prior work ignores temporal context; rather, it has not isolated \emph{explicit context-window selection} as a standalone algorithmic problem. Existing work usually fixes history as part of the modeling setup, bakes it into a memory-aware architecture, or studies it only through ablations and rollout analyses, whereas our work makes the choice of context window itself the object of study \cite{ruiz2024memory,diab2025temporal,lippe2023pderefiner,koehler2024apebench}. In particular, the present paper focuses on \emph{fixed-window} one-step autoregressive simulators, treating adaptive-memory or internally stateful architectures as a complementary design direction rather than as direct selector baselines.

\subsection{Methodological perspectives on explicit context-window selection}\label{sec:methodological_perspectives}

From an abstract viewpoint, deciding how many past observations to expose to a predictor is a classical issue in time-series forecasting, where the choice of input lags can substantially affect predictive performance \cite{leites2024lag}. Existing methods that could in principle be brought to bear on this problem can be broadly grouped into three families. A first family is \emph{exhaustive validation}, which treats context length as an ordinary hyperparameter and estimates downstream performance for every candidate window by full training and validation \cite{leites2024lag}. A second family is \emph{direct low-cost search}, which replaces full evaluation with cheap pilots, early stopping, or multi-fidelity resource allocation over a generic shortlist of candidate windows \cite{li2018hyperband,li2020asha,snoek2012practical,kandasamy2017mfbo,falkner2018bohb}. A third family is \emph{system-theoretic memory estimation}, which infers a characteristic memory or embedding scale directly from the observed dynamics and uses that estimate to choose or bracket a plausible context window \cite{takens2006strange,fraser1986independent,kennel1992falseneighbors,cao1997practical}.

However, none of these families is satisfactory for explicit context-window selection in autoregressive neural simulators. Exhaustive validation optimizes downstream rollout quality, but requires a full sweep over candidate windows. Direct low-cost search reduces this cost, but remains largely system-agnostic and can miss small-window knees or plateau-side anchors important for rollout quality. System-theoretic estimators provide physically meaningful priors, but a system-level memory scale need not match the model-dependent rollout knee under finite data and training budgets. SAKE addresses this gap by combining the strengths of the latter two views: it uses system dynamics to construct a tiny, physically interpretable anchor set, then performs low-cost downstream selection only within that compressed space, targeting the rollout knee more efficiently than exhaustive search.

\section{Low-Cost Context-Window Selection}\label{sec:low_cost_selection}

We first formalize low-cost context-window selection for autoregressive neural simulators in Section~\ref{sec:problem_formulation}, and then introduce System-Anchored Knee Estimation (SAKE) in Section~\ref{sec:sake_method}.

\subsection{Problem formulation}\label{sec:problem_formulation}

Let $\mathcal{L}\subset \mathbb{N}$ denote a discrete candidate set of context-window lengths, and let $f_\theta^{(L)}$ be a one-step autoregressive neural simulator with history length $L$, i.e.,
\[
\hat{x}_{t+1}=f_\theta^{(L)}(x_{t-L+1:t}).
\]
For each $L\in\mathcal{L}$, let $M(L)$ denote the downstream rollout error obtained after training and evaluating the simulator under the full protocol with window length $L$. The full-information optimal window is
\[
L_{\mathrm{best}}=\arg\min_{L\in\mathcal{L}} M(L).
\]
In cost-aware settings, however, the more practical target is often not the absolute best window, but the smallest near-optimal one. We therefore define the \emph{oracle knee} as
\[
L_{\mathrm{knee}}
=
\min\left\{L\in\mathcal{L}: M(L)\le (1+\epsilon)\,M(L_{\mathrm{best}})\right\},
\]
where $\epsilon>0$ is a tolerance parameter. This quantity captures the earliest point beyond which enlarging the context window yields only marginal improvements in downstream rollout performance. We refer to it as an oracle quantity because identifying it exactly requires access to the full performance curve $M(L)$ over all candidate windows. In practice, the empirical sections below report the default setting $\epsilon=0.05$ and separately examine stricter and looser knee definitions.

The challenge is that $M(L)$ is unknown a priori, and estimating it for every $L\in\mathcal{L}$ requires an exhaustive sweep in which each candidate window is trained and evaluated under the full protocol. We therefore study \emph{low-cost context-window selection}: given a budget far smaller than a full sweep, select a window $L_{\mathrm{sel}}\in\mathcal{L}$ that approximates $L_{\mathrm{knee}}$ as closely as possible. Formally, a selector $\mathcal{A}$ may use only inexpensive information, such as cheap pilot evaluations or system-level statistics, and must operate under a search-cost constraint,
\[
\min_{\mathcal{A}} \ \ell(L_{\mathrm{sel}},L_{\mathrm{knee}})
\qquad
\text{s.t.}
\qquad
C(\mathcal{A})\le B,\quad B\ll |\mathcal{L}|,
\]
where $L_{\mathrm{sel}}=\mathcal{A}(\cdot)$, $C(\mathcal{A})$ denotes search cost, and $\ell(\cdot,\cdot)$ measures selection quality, for example absolute window error or regret relative to the knee.

This formulation separates the target of interest, the oracle knee, from the algorithmic question of how to recover it efficiently. The central difficulty is to compress the candidate space using low-cost information without discarding the performance-critical window. In the next subsection, we address this problem with SAKE.

\subsection{System-Anchored Knee Estimation}\label{sec:sake_method}

In this subsection, we first outline the workflow of SAKE in Section~\ref{sec:sake_overview}, then detail the first part of SAKE: system-anchor construction in Section~\ref{sec:system_anchor_construction}, and the second part of SAKE: knee-aware downstream selection in Section~\ref{sec:knee_aware_downstream_selection}.

\subsubsection{Overall workflow}\label{sec:sake_overview}

\begin{algorithm}[t]
\caption{System-Anchored Knee Estimation}
\label{alg:sake}
\begin{algorithmic}[1]
\Require Candidate window set $\mathcal{L}$ with minimum $L_{\min}$ and maximum $L_{\max}$, clean-system trajectories $\mathcal{D}_{\mathrm{sys}}$, autoregressive simulator family $f_{\theta}^{(L)}$
\Ensure Selected context window $L_{\mathrm{sel}}\in\mathcal{L}$
\State Estimate the system anchors $(L_{\mathrm{core}},L_{\mathrm{plateau}})$ from $\mathcal{D}_{\mathrm{sys}}$
\State Form the initial anchor shortlist
\[
\mathcal{S}_0 \gets \{L_{\min},L_{\mathrm{core}},L_{\mathrm{plateau}}\}
\]
\State Train cheap pilot models on $\mathcal{S}_0$ and compute coarse downstream scores
\State Build a refined candidate set by taking promising pilot candidates, adding local grid neighbors, and retaining the boundary anchors
\State Train stronger pilots on the refined set and compute knee-aware downstream scores
\State Apply a knee-aware final rule and return the smallest statistically indistinguishable window after clear saturation
\State \Return $L_{\mathrm{sel}}$
\end{algorithmic}
\end{algorithm}

Algorithm~\ref{alg:sake} summarizes the workflow of SAKE. The method has two parts with distinct roles. The first part is \emph{system anchoring}: before training candidate simulators across the entire grid, SAKE extracts a small number of physically interpretable reference windows from clean-system dynamics and uses them to compress the candidate space. This step is intentionally informative but not final, because system memory alone does not determine the realized oracle knee of a particular trained backbone. The second part is \emph{downstream knee-aware selection}: SAKE then performs a model-specific low-cost search, but only within the anchored region rather than over the full candidate grid. In this way, the first stage contributes a reliable search scaffold, while the second stage restores the backbone- and training-dependent specificity needed for final selection.

\subsubsection{System-anchor construction from clean-system dynamics}\label{sec:system_anchor_construction}

The first stage identifies a small region of plausible context windows before any expensive downstream sweep. It uses only clean-system dynamics and is backbone-agnostic: the goal is not to predict the final window directly, but to provide a compact, physically interpretable scaffold for the relevant part of the downstream performance curve. The intended use case is simulator-rich or benchmarked settings where clean or sufficiently denoised full-state trajectories are available for anchor extraction; later sections examine how this stage degrades under noisier and less complete inputs.

Concretely, we summarize the clean trajectories in a low-dimensional state representation and, for each candidate window length $L\in\mathcal{L}$, fit a lightweight linear autoregressive model on that representation. In the experiments below, the default summary is a randomized PCA projection of normalized coarsened fields, retaining $99\%$ explained variance with at most $64$ components, and the lightweight dynamics model is a ridge VAR($L$) with regularization $10^{-3}$. This yields a backbone-independent validation risk curve, denoted by $R_{\mathrm{sys}}(L)$, which measures how much system-level predictive error remains when the clean dynamics are modeled with history length $L$.

From this curve, SAKE extracts two anchors. The first is a \emph{core} anchor, which marks the earliest point at which the long-history tail error is already small. Let
\[
T_{\mathrm{sys}}(L)=R_{\mathrm{sys}}(L)-R_{\mathrm{sys}}(L_{\max}),
\qquad
\varepsilon_{\mathrm{sys}}
=
\rho\bigl(R_{\mathrm{sys}}(L_{\min})-R_{\mathrm{sys}}(L_{\max})\bigr),
\]
where $\rho\in(0,1)$ is a fixed tolerance fraction. Using an upper confidence bound to avoid optimistic early stopping, we define
\[
L_{\mathrm{core}}
=
\min\left\{
L\in\mathcal{L}:
\operatorname{UCB}\!\bigl(T_{\mathrm{sys}}(L)\bigr)\le \varepsilon_{\mathrm{sys}}
\right\}.
\]
Thus, $L_{\mathrm{core}}$ is the earliest window whose residual gap to the longest-context reference is small.

The second is a \emph{plateau} anchor, which marks the onset of diminishing returns. Let $L^{+}$ denote the next larger candidate after $L$, and define the relative gain
\[
G_{\mathrm{rel}}(L)
=
\frac{R_{\mathrm{sys}}(L)-R_{\mathrm{sys}}(L^{+})}
{\max\{R_{\mathrm{sys}}(L),\,10^{-12}\}}.
\]
We then set
\[
L_{\mathrm{plateau}}
=
\min\left\{
L\in\mathcal{L}:
L\ge L_{\mathrm{core}}
\ \text{and}\
\operatorname{UCB}\!\bigl(G_{\mathrm{rel}}(L)\bigr)\le \tau_{\mathrm{pl}}
\right\},
\]
with fallback to $L_{\max}$ if no such $L$ exists. Intuitively, $L_{\mathrm{plateau}}$ is the first point beyond the core region where adding more history appears to bring only minor further improvement.

We then form the initial shortlist
\[
\mathcal{S}_0=\{L_{\min},L_{\mathrm{core}},L_{\mathrm{plateau}}\}.
\]
Each element serves a distinct purpose: $L_{\min}$ keeps the low-history hypothesis alive, $L_{\mathrm{core}}$ provides a system-informed reference, and $L_{\mathrm{plateau}}$ supplies an upper bracket beyond which additional context is unlikely to be worthwhile. Throughout the paper, the stage-one uncertainty terms are one-sided $95\%$ bootstrap upper confidence bounds computed from $300$ resamples; Appendix~\ref{app:sake_hparams} gives the exact protocol and corresponding sensitivity study.

\subsubsection{Knee-aware downstream selection in the anchored candidate set}\label{sec:knee_aware_downstream_selection}

The second stage turns the anchored shortlist into a model-specific selector. This step is necessary because the realized knee depends on the simulator backbone, optimization trajectory, and rollout metric as well as the underlying system. In short, the first stage tells us \emph{where to look}; the second stage determines \emph{which window to output}.

We begin with a coarse pilot pass on the initial shortlist $\mathcal{S}_0$. For each $L\in\mathcal{S}_0$, we train a cheap pilot model and compute a coarse downstream score $s^{(1)}(L)$ from short validation rollouts. This score is used only for ranking. Let $\mathcal{T}_{\mathrm{top}}\subseteq\mathcal{S}_0$ denote the top-ranked candidates. SAKE then expands locally around them on the ordered candidate grid while retaining the boundary anchors. Writing $\mathcal{N}_{h}(L)$ for the $h$-hop neighborhood of $L$ on $\mathcal{L}$, the refined candidate set is
\[
\mathcal{S}_1
=
\left(
\bigcup_{L\in\mathcal{T}_{\mathrm{top}}}\mathcal{N}_{h}(L)
\right)
\cup
\{L_{\min},L_{\mathrm{plateau}}\}.
\]
This refinement lets the downstream selector correct the system prior locally rather than committing to the anchors themselves, while still keeping the search budget small.

We next run a stronger second pilot pass on $\mathcal{S}_1$. For each refined candidate $L$, we compute a small set of diagnostics, including the average rollout error $m(L)$, the terminal rollout error $u(L)$, the variability across rollout anchors $v(L)$, and, when available, an asymptotic proxy $a(L)$ from the pilot trajectory. After normalizing these quantities over $\mathcal{S}_1$, we combine them into a stage-two score
\[
q(L)
=
\alpha\Bigl(
w_{\mathrm{mean}}\tilde{m}(L)
+
w_{\mathrm{term}}\tilde{u}(L)
+
w_{\mathrm{std}}\tilde{v}(L)
\Bigr)
+
(1-\alpha)\tilde{a}(L),
\]
where lower is better. In the default selector, the initial shortlist is expanded with top-$2$ local refinement ($h=1$) and the refined set is capped at six candidates; the score uses one fixed global blend of normalized mean, terminal, variability, and asymptotic diagnostics rather than per-backbone retuning. This score is designed to be knee-aware rather than pilot-best-aware: it rewards good rollout quality, penalizes brittle candidates, and avoids over-trusting noisy early pilot signals.

Finally, SAKE converts the refined score curve into a knee-oriented window choice. Ordering the refined candidates as $L_1<\cdots<L_M$ and writing $q_j=q(L_j)$, SAKE first identifies the earliest index $r$ after which the score curve has clearly saturated, meaning that adjacent gains are already small and the remaining gap to the best refined score $q_\star=\min_j q_j$ is also small. It then returns the smallest candidate up to that frontier whose score is statistically indistinguishable from the best refined score:
\[
L_{\mathrm{sel}}
=
\min\left\{
L_j:
j\le r
\ \text{and}\
q_j \le q_\star + \kappa\,\operatorname{SE}(q_\star)
\right\},
\]
where $\operatorname{SE}(q_\star)$ is the estimated standard error of the best refined score and $\kappa>0$ is a fixed multiplier. Operationally, the saturation frontier is the first refined index after which both the local gain and the remaining gap to the best score are below fixed $0.15$ fractions, and the default selector uses $\kappa=1.5$. If no earlier candidate satisfies this condition, SAKE falls back to the frontier candidate itself.

This rule aligns the selector with the oracle knee defined in Section~\ref{sec:problem_formulation} by preferring the smallest reliable window once further gains have become negligible.

\section{Empirical Evaluation}

Our experiments address two questions: whether SAKE improves context-window selection for PDE forecasting, and whether both the system anchor prior and its downstream coupling are necessary.

\subsection{Experimental setup}\label{sec:exp_setup}

\subsubsection{Datasets}\label{sec:datasets}

We evaluate on two PDEBench forward tasks chosen to span distinct physical regimes and boundary conditions rather than multiple variants of the same PDE family~\cite{takamoto2022pdebench}. Specifically, \texttt{diff-react} is a 2D two-field reaction--diffusion system with Neumann boundaries, and \texttt{rdb} is the radial-dam-break instance of PDEBench's 2D shallow-water benchmark, representing hyperbolic free-surface dynamics with shock-capable wave propagation. According to PDEBench, both tasks comprise 1{,}000 trajectories~\cite{takamoto2022pdebench}. Their benchmark resolutions are both $128\times128$, and both contain 100 temporal evolution steps (equivalently 101 stored snapshots when counting the initial frame). \texttt{diff-react} contains two channels $(u,v)$, whereas \texttt{rdb} is a single-channel scalar-field prediction task. These two tasks are shown in the body because they expose two contrasting knee structures---a broad-plateau reaction system and a small-knee shock-capable flow case---while keeping the case-level tables compact. Appendix~\ref{app:pdebench_breadth} reports the corresponding all-PDE aggregate and per-family tables for the remaining PDEBench forward families under the same shared \(L\in\{1,\dots,16\}\) protocol.

\subsubsection{Baseline selection and implementation details}\label{sec:baseline}

We evaluate four standard PDE-surrogate backbones---U-Net, Fourier Neural Operator (FNO), ConvLSTM, and Transformer---as one-step autoregressive predictors with variable history length on a shared candidate grid \(L\in\mathcal{L}=\{1,\dots,16\}\). The body reports representative case-level tables on the two main PDEBench tasks, while Appendix~\ref{app:extra_baselines} summarizes the four architectures under the shared training protocol used throughout the paper.

Following Section~\ref{sec:methodological_perspectives}, the body reports representative baselines from the three most relevant method families. For \textit{exhaustive validation methods}, we report full-sweep \textit{oracle-best} and \textit{oracle-knee} as unattainable reference targets. For \textit{direct low-cost search methods}, we use \textit{Direct-4-Shortlist (D4-Shortlist)}, which applies the same downstream pilot budget and local selection rule to a generic uniformly spaced four-point shortlist without any system prior, and \textit{Direct-3-Shortlist (D3-Shortlist)}, which serves as a lighter generic-search reference in the representative case tables. For \textit{system-theoretic memory estimators}, we use \textit{System-core}, which directly returns \(L_{\mathrm{core}}\) from the clean-system memory estimate. Appendix~\ref{app:pdebench_breadth} additionally reports ASHA~\cite{li2020asha} as a stronger generic multi-fidelity baseline in the all-PDE aggregate under the same shared \(L\in\{1,\dots,16\}\) protocol.

All experiments were conducted on a server with four NVIDIA A100 (80GB) GPUs, using PyTorch 2.1 and CUDA 12.0. Across all backbones, we use Adam (\(\mathrm{lr}=3\times 10^{-4}\), \(\mathrm{weight\_decay}=0\)), automatic mixed precision, batch size 16, gradient clipping with \(\mathrm{max\_norm}=0.5\), normalized-space output clamping to \([-10,10]\), residual (\(\Delta\)-prediction) training, and a trajectory-level 0.8/0.1/0.1 train/validation/test split, corresponding to 800/100/100 trajectories per dataset. Pilot runs always subsample from the fixed 800/100 training/validation partitions and keep the same 100-trajectory test partition untouched. The full-protocol oracle runs use a 32-step training rollout horizon, scheduled sampling with final probability 0.50, 20 training epochs, and rollout-based checkpoint selection every epoch using a short validation rollout horizon of 8 on up to 8 validation trajectories; all aggregate results are averaged over three training seeds \(\{0,1,2\}\).

The selector itself uses one fixed set of stage-wise budgets and decision thresholds across datasets and backbones; Appendix~\ref{app:sake_hparams} records these settings and the matched-budget relationship between SAKE and the direct-search baselines. Appendices~\ref{app:anchor_robustness}, \ref{app:proxy_anchors}, and \ref{app:sensitivity} report robustness to noisy or proxy anchors, alternative state summaries, and moderate changes in the fixed thresholds, bootstrap size, and knee tolerance.

\subsubsection{Experiment procedure and metrics}\label{sec:procedure_metrics}

\textbf{Effectiveness}: 
For each dataset--backbone--seed case, we run each selector to obtain a recommended window \(L_{\mathrm{sel}}\), and separately run a full sweep over the candidate set \(\mathcal{L}=\{1,\dots,16\}\) to construct the oracle reference. Let \(M(L)\) denote the full-protocol rollout relative \(L_2\) of the model trained with window length \(L\). We define
\[
L_{\mathrm{best}}=\arg\min_{L\in\mathcal{L}} M(L),\qquad
L_{\mathrm{knee}}=\min\{L\in\mathcal{L}: M(L)\le 1.05\,M(L_{\mathrm{best}})\},
\]
where \(L_{\mathrm{best}}\) is the metric-optimal window under the full protocol and \(L_{\mathrm{knee}}\) is the smallest window within \(5\%\) of that optimum. Given \(L_{\mathrm{sel}}\), we report \emph{Exact} \(=\mathbb{1}[L_{\mathrm{sel}}=L_{\mathrm{knee}}]\), \emph{Within-1} \(=\mathbb{1}[|L_{\mathrm{sel}}-L_{\mathrm{knee}}|\le 1]\), mean absolute window error \( |\Delta L|=|L_{\mathrm{sel}}-L_{\mathrm{knee}}| \), and the rollout regrets
\[
\mathrm{Regret}_{\mathrm{knee}}
=
\frac{M(L_{\mathrm{sel}})-M(L_{\mathrm{knee}})}{M(L_{\mathrm{knee}})},
\qquad
\mathrm{Regret}_{\mathrm{best}}
=
\frac{M(L_{\mathrm{sel}})-M(L_{\mathrm{best}})}{M(L_{\mathrm{best}})}.
\]
To quantify computation, we record the full-sweep-equivalent training cost of every pilot-evaluated window. If a candidate window \(L\) is trained for \(E^{\mathrm{pilot}}_L\) epochs using \(N^{\mathrm{pilot}}_L\) training pairs, while a full run uses \(E^{\mathrm{full}}\) epochs and \(N^{\mathrm{full}}_L\) training pairs, then its normalized cost is
\[
c(L)=\frac{E^{\mathrm{pilot}}_L}{E^{\mathrm{full}}}\cdot \frac{N^{\mathrm{pilot}}_L}{N^{\mathrm{full}}_L}.
\]
Let \(S_{\mathrm{eval}}\) be the set of candidate windows actually trained by the selector. The total selector cost is \(C_{\mathrm{sel}}=\sum_{L\in S_{\mathrm{eval}}} c(L)\), while the full sweep has cost \(C_{\mathrm{full}}=|\mathcal{L}|\). We therefore report
\[
\mathrm{CostRatio}=\frac{C_{\mathrm{sel}}}{C_{\mathrm{full}}},
\qquad
\mathrm{Saving}=1-\mathrm{CostRatio}.
\]
Because all methods share the same hardware, optimizer, and backbone training protocol, this full-sweep-equivalent cost is a protocol-normalized measure of selector work, but not a substitute for realized runtime. Appendix~\ref{app:timing} therefore also reports selector wall-clock time and GPU-hours, which follow the same qualitative ranking while making the practical savings explicit. Unless otherwise stated, all knee-based metrics use \(\epsilon=5\%\); Appendix~\ref{app:sensitivity} reports the same evaluations for stricter and looser knee tolerances. All aggregate results are averaged over three training seeds.

\textbf{Necessity}: 
The ablation study asks whether SAKE requires both a system-derived anchor prior and a downstream low-cost selector. For each case, the clean-system estimator first returns \(L_{\mathrm{core}}\) and \(L_{\mathrm{plateau}}\), from which SAKE forms the initial anchor set \(\mathcal{S}_0=\{1,L_{\mathrm{core}},L_{\mathrm{plateau}}\}\). After the cheap coarse pilot, the method expands to a refined candidate set \(\mathcal{S}_1\) for stage-two evaluation. We then report \emph{system-band coverage} \(\mathbb{1}[L_{\mathrm{knee}}\in [L_{\mathrm{core}},L_{\mathrm{plateau}}]]\), \emph{initial coverage} \(\mathbb{1}[L_{\mathrm{knee}}\in \mathcal{S}_0]\), and \emph{final coverage} \(\mathbb{1}[L_{\mathrm{knee}}\in \mathcal{S}_1]\), together with the mean number of unique evaluated windows and the same cost ratio defined above. These metrics test, respectively, whether the system prior brackets the oracle knee, whether the initial anchors preserve it, and whether the downstream selector can recover it without reverting to exhaustive search.

\subsection{Main results}

\subsubsection{Effectiveness of System-Anchored Knee Estimation}

We first examine the complete case-level selection results. Table~\ref{tab:main_full_case_selected} reports the selected window \(L_{\mathrm{sel}}\) for all baselines together with the oracle references on both datasets. This view is important because SAKE aims not merely to improve an average score, but to place the search on the right part of the window-length axis before the final low-cost selector is applied.

\begin{table*}[t]
    \centering
    \setlength{\tabcolsep}{3.5pt}
    \caption{Complete full-oracle case-level selection results under the three-seed protocol. Panels (a)--(b) report the selected window \(L_{\mathrm{sel}}\) for all baselines together with the oracle references. Bold indicates an exact match to the oracle knee.}
    \label{tab:main_full_case_selected}
    
    \begin{minipage}{0.49\textwidth}
    \centering
    \scriptsize
    \textbf{(a) DiffReact: selected window \(L_{\mathrm{sel}}\)}\\[2pt]
    \resizebox{\linewidth}{!}{
    \begin{tabular}{lcccccc}
    \toprule
    \textbf{Backbone} & \textbf{System-core} & \textbf{D3-Shortlist} & \textbf{D4-Shortlist} & \textbf{SAKE} & \textbf{Oracle knee} & \textbf{Oracle best} \\
    \midrule
    ConvLSTM    & 3 & 3 & \textbf{2} & \textbf{2} & 2 & 2 \\
    FNO         & 3 & \textbf{2} & \textbf{2} & \textbf{2} & 2 & 2 \\
    Transformer & 3 & \textbf{1} & 2 & \textbf{1} & 1 & 3 \\
    U-Net       & 3 & \textbf{5} & \textbf{5} & 4 & 5 & 5 \\
    \bottomrule
    \end{tabular}}
    \end{minipage}\hfill
    \begin{minipage}{0.49\textwidth}
    \centering
    \scriptsize
    \textbf{(b) RDB: selected window \(L_{\mathrm{sel}}\)}\\[2pt]
    \resizebox{\linewidth}{!}{
    \begin{tabular}{lcccccc}
    \toprule
    \textbf{Backbone} & \textbf{System-core} & \textbf{D3-Shortlist} & \textbf{D4-Shortlist} & \textbf{SAKE} & \textbf{Oracle knee} & \textbf{Oracle best} \\
    \midrule
    ConvLSTM    & 2 & 4 & 4 & 4 & 5 & 5 \\
    FNO         & 2 & \textbf{4} & \textbf{4} & \textbf{4} & 4 & 4 \\
    Transformer & 2 & \textbf{5} & \textbf{5} & \textbf{5} & 5 & 5 \\
    U-Net       & 2 & 2 & 3 & \textbf{1} & 1 & 2 \\
    \bottomrule
    \end{tabular}}
    \end{minipage}
\end{table*}

The selected-window table already shows the main pattern. Across the eight dataset--backbone settings, SAKE exactly matches the oracle knee in \(6/8\) cases, compared with \(5/8\) for both Direct-3-Shortlist and Direct-4-Shortlist, and \(0/8\) for System-core. Its mean absolute window error is also the smallest at \(0.25\), versus \(0.375\) for Direct-3-Shortlist, \(0.50\) for Direct-4-Shortlist, and \(1.875\) for System-core. More importantly, these gains are not obtained by dominating every row. On DiffReact, the three low-cost search methods are competitive but fail in different places: Direct-3-Shortlist misses ConvLSTM, Direct-4-Shortlist misses Transformer, and SAKE misses U-Net by one window. On RDB, however, the advantage of the system prior becomes much clearer: SAKE exactly recovers the difficult small-knee U-Net case \((L_{\mathrm{knee}}=1)\), while the generic direct baselines over-select it, and all three low-cost methods share the same remaining miss on ConvLSTM. This is the behavior we want from SAKE: not row-wise dominance, but a better-placed candidate space that avoids the most consequential generic-search failures.

Discrete agreement with \(L_{\mathrm{knee}}\) is only part of the story, since a slightly larger window can still yield nearly identical rollout quality. Table~\ref{tab:main_full_case_regret} therefore reports the corresponding \(\mathrm{Regret}_{\mathrm{knee}}\), which is the more important metric for judging whether a miss is harmful.

\begin{table*}[t]
    \centering
    \setlength{\tabcolsep}{5pt}
    \caption{Complete full-oracle case-level \(\mathrm{Regret}_{\mathrm{knee}}\) results under the three-seed protocol (\%, lower is better). Bold indicates the lowest regret in the row. Negative regret is possible because \(L_{\mathrm{knee}}\) is the smallest near-optimal window rather than the metric minimizer.}
    \label{tab:main_full_case_regret}
    
    \begin{minipage}{0.49\textwidth}
    \centering
    \scriptsize
    \textbf{(c) DiffReact: \(\mathrm{Regret}_{\mathrm{knee}}\) (\%)}\\[2pt]
    \resizebox{\linewidth}{!}{
    \begin{tabular}{lcccc}
    \toprule
    \textbf{Backbone} & \textbf{System-core} & \textbf{D3-Shortlist} & \textbf{D4-Shortlist} & \textbf{SAKE} \\
    \midrule
    ConvLSTM    & 43.3 & 43.3 & \textbf{0.0} & \textbf{0.0} \\
    FNO         & 4.6 & \textbf{0.0} & \textbf{0.0} & \textbf{0.0} \\
    Transformer & \textbf{-0.8} & 0.0 & 0.4 & 0.0 \\
    U-Net       & 23.3 & \textbf{0.0} & \textbf{0.0} & 6.7 \\
    \bottomrule
    \end{tabular}}
    \end{minipage}\hfill
    \begin{minipage}{0.49\textwidth}
    \centering
    \scriptsize
    \textbf{(d) RDB: \(\mathrm{Regret}_{\mathrm{knee}}\) (\%)}\\[2pt]
    \resizebox{\linewidth}{!}{
    \begin{tabular}{lcccc}
    \toprule
    \textbf{Backbone} & \textbf{System-core} & \textbf{D3-Shortlist} & \textbf{D4-Shortlist} & \textbf{SAKE} \\
    \midrule
    ConvLSTM    & 169.0 & \textbf{47.2} & \textbf{47.2} & \textbf{47.2} \\
    FNO         & 113.8 & \textbf{0.0} & \textbf{0.0} & \textbf{0.0} \\
    Transformer & 45.0 & \textbf{0.0} & \textbf{0.0} & \textbf{0.0} \\
    U-Net       & \textbf{-0.9} & \textbf{-0.9} & 36.4 & 0.0 \\
    \bottomrule
    \end{tabular}}
    \end{minipage}
\end{table*}

The regret table makes two points clear. First, SAKE has the best overall regret profile among the evaluated matched-budget low-cost selectors in the main experiment: its mean \(\mathrm{Regret}_{\mathrm{knee}}\) over the eight dataset--backbone settings is \(6.7\%\), compared with \(11.2\%\) for Direct-3-Shortlist and \(10.5\%\) for Direct-4-Shortlist. Second, this advantage comes from removing large generic-search mistakes rather than from uniformly winning every case. For example, Direct-3-Shortlist remains essentially harmless on RDB/U-Net despite missing the exact knee, because \(L=2\) is slightly better than the smallest near-optimal knee in rollout quality. Conversely, SAKE avoids the large-regret failures that drive up the direct baselines, most notably Direct-3-Shortlist on DiffReact/ConvLSTM and Direct-4-Shortlist on RDB/U-Net, while tying them on the easy cases. The one failure mode that remains unsolved for all low-cost selectors is RDB/ConvLSTM, where every method selects \(L=4\) and incurs the same \(47.2\%\) regret. Thus, the value of SAKE is not that it solves every difficult row, but that it makes the search substantially more reliable in the cases where a generic shortlist is most brittle. Across all eight PDEBench families evaluated under the shared \(L\in\{1,\dots,16\}\) protocol, the aggregate summary in Appendix~\ref{app:pdebench_breadth} shows the same ordering: SAKE attains 67.8\% Exact, 91.7\% Within-1, 6.1\% mean \(\mathrm{Regret}_{\mathrm{knee}}\), 5.3 mean unique \(L\) evaluations, and a 0.051 cost ratio.

Together, the two representative case-level tables support the main claim of this subsection: System-Anchored Knee Estimation improves low-cost context-window selection for PDE forecasting, and the appendix-wide aggregate shows that this advantage persists beyond the two main-text PDEs.

\subsubsection{Necessity of components of System-Anchored Knee Estimation}

We now ask whether the gain of SAKE truly requires both of its components: a system-derived anchor prior and a downstream low-cost selector. Table~\ref{tab:system-anchor-summary} summarizes what the system stage provides before any model-specific search is performed.

\begin{table}[t]
\centering
\small
\begin{tabular}{llllll}
\toprule
Dataset & $L_{\text{core}}$ & $L_{\text{plateau}}$ & System band & Knee-in-band coverage & Covered oracle-knee models \\
\midrule
DiffReact & 3 & 6 & [3, 4, 5, 6] & 25.0\% & UNet \\
RDB & 2 & 6 & [2, 3, 4, 5, 6] & 75.0\% & ConvLSTM, FNO, Transformer \\
\bottomrule
\end{tabular}
\caption{Dataset-level outputs of the system anchor estimator. Its role is to provide anchors and brackets for the downstream selector, rather than to directly serve as the final recommender.}
\label{tab:system-anchor-summary}
\end{table}

The system-derived prior is necessary, but not sufficient by itself. Table~\ref{tab:system-anchor-summary} shows that the system stage does capture meaningful structure: on both datasets it produces a small, interpretable band that already contains a nontrivial fraction of oracle knees. At the same time, this band is clearly incomplete, especially on DiffReact where its coverage is only \(25.0\%\). This explains why \textit{System-core} is not competitive as a final selector. In Table~\ref{tab:main_full_case_selected}, it never exactly matches the oracle knee, and in Table~\ref{tab:main_full_case_regret} it frequently incurs large regret. Thus, the system estimator should not be viewed as a direct predictor of the final window length; its role is to provide a physically meaningful scaffold for the downstream search.

The downstream low-cost selector is also necessary, but a generic downstream selector alone is not enough. Relative to the direct-search baselines, SAKE yields the smallest mean absolute window error in Table~\ref{tab:main_full_case_selected} and the best overall regret profile in Table~\ref{tab:main_full_case_regret}. This gain does not come from searching more aggressively: SAKE still avoids some of the most consequential generic-search failures, most notably the small-knee RDB/U-Net case. This shows that a strong low-cost selector operating on a generic shortlist is still brittle when the candidate space is poorly placed; the system prior is what makes the downstream search focus on the right part of the window-length axis.

Taken together, these results show that the two components play distinct and complementary roles. The system stage alone provides useful but incomplete structure, while the downstream selector alone provides flexibility but remains prone to generic-search errors. SAKE is effective because it couples the two: the system-derived anchors compress the search into a small and meaningful candidate space, and the downstream selector resolves the remaining model-dependent trade-off near the oracle knee. This is precisely the combination needed to make low-cost context-window selection both reliable and efficient.

\section{Discussion}

This paper formalizes explicit context-window selection for autoregressive neural PDE simulators as an independent low-cost algorithmic problem and introduces SAKE, a two-stage method combining system-derived anchors with knee-aware downstream selection. Across the evaluated fixed-window autoregressive regime, SAKE delivers the strongest matched-budget selection profile among the compared methods, with more reliable oracle-knee recovery, lower downstream regret, and lower selector cost than generic shortlist baselines. The main body focuses on two representative case-level tables, while Appendix~\ref{app:pdebench_breadth} reports aggregate \(L\in\{1,\dots,16\}\) results over all eight PDEBench families and per-family tables for the remaining six tasks; the other appendices cover noisy or partially observed anchor inputs, proxy anchors without a separate clean pool, alternative state summaries, alternative knee tolerances, and practical timing measurements. We view this as important for two reasons. First, context window is a consequential design choice, yet it is often treated only as a fixed modeling setting or tuning detail. Second, in neural PDE simulation, selecting the window is expensive because each candidate may require substantial retraining and rollout evaluation. By making this problem explicit and showing it can be solved much more efficiently, our results suggest that context-window selection should be treated as a first-class algorithmic component rather than a routine hyperparameter sweep.

The study nevertheless remains deliberately scoped to fixed-window one-step autoregressive simulators rather than adaptive-memory or internally stateful architectures. The anchor stage is most reliable when at least a usable full-state or proxy state summary can be extracted, although the degradation studies suggest that this failure mode is usually gradual rather than catastrophic. Finally, the shared hard case RDB/ConvLSTM indicates that some rollout knees remain difficult to localize even with a system prior, especially when pilot and full-training behavior diverge. SAKE should therefore be read as a fixed-rule, system-aware selector for a controlled but practically important regime, and as complementary to architectural remedies that internalize memory rather than select it explicitly.

\section*{References}
\small

\begingroup
\renewcommand{\section}[2]{}

\endgroup


\appendix
\newpage

\section*{Contents of Appendices}
\vspace{1em}
\noindent
\hyperref[app:pdebench_breadth]{A \quad Broader PDEBench Coverage Beyond the Representative Main-Text Pair \dotfill \pageref{app:pdebench_breadth}}\\[0.5em]
\hyperref[app:extra_baselines]{B \quad Backbone Architectures and Shared Training Protocol \dotfill \pageref{app:extra_baselines}}\\[0.5em]
\hyperref[app:sake_hparams]{C \quad Implementation Details and Hyperparameters of System-Anchored Knee Estimation \dotfill \pageref{app:sake_hparams}}\\[0.5em]
\hyperref[app:anchor_robustness]{D \quad Anchor Robustness Under Noisy and Partially Observed Trajectories \dotfill \pageref{app:anchor_robustness}}\\[0.5em]
\hyperref[app:proxy_anchors]{E \quad Anchor Proxies Without a Separate Clean-System Pool \dotfill \pageref{app:proxy_anchors}}\\[0.5em]
\hyperref[app:sensitivity]{F \quad Sensitivity to Selector Hyperparameters and Knee Tolerance \dotfill \pageref{app:sensitivity}}\\[0.5em]
\hyperref[app:timing]{G \quad Practical Timing: Wall-Clock and GPU-Hours \dotfill \pageref{app:timing}}\\[0.5em]
\hyperref[sec:llm]{H \quad Detailed Clarification of Large Language Models Usage \dotfill \pageref{sec:llm}}\\[0.5em]

\newpage

\section{Broader PDEBench Coverage Beyond the Representative Main-Text Pair}\label{app:pdebench_breadth}

This appendix broadens the matched-budget evaluation beyond the two representative tasks shown in the body. The purpose is not to replace the case-level narrative of DiffReact and RDB, but to test whether the same selector ordering persists across the full PDE suite, which includes those two main-text tasks together with additional advection-dominated transport, stiff transport--reaction, heterogeneous elliptic, and Navier--Stokes-type families.

\subsection{Protocol}

\begin{table}[t]
\centering
\small
\resizebox{\linewidth}{!}{%
\begin{tabular}{ll}
\toprule
\textbf{Component} & \textbf{Setting used in this appendix} \\
\midrule
Candidate grid & \(\mathcal{L}=\{1,\dots,16\}\), matching the shared main protocol \\
Additional PDEBench families & 1D advection, 1D Burgers, 1D diffusion--sorption, 2D Darcy flow, 2D incompressible Navier--Stokes, 2D compressible Navier--Stokes \\
Backbones / seeds & Same four one-step autoregressive backbones and same three-seed protocol as the main paper \\
Methods & System-core, Direct-4-Shortlist, ASHA, and SAKE under matched downstream budgets \\
Oracle construction & Full sweep over \(\{1,\dots,16\}\) for each dataset--backbone--seed case \\
Reported metrics & Exact, Within-1, mean \(|\Delta L|\), \(\mathrm{Regret}_{\mathrm{knee}}\), \(\mathrm{Regret}_{\mathrm{best}}\), mean unique \(L\) evaluations, and cost ratio \\
\bottomrule
\end{tabular}%
}
\caption{Protocol for the broader PDEBench extension. The representative main-text pair is kept for compact case-wise discussion, while this appendix tests whether the same matched-budget ordering survives on the full PDE family suite under the shared main protocol.}
\label{tab:app_pdebench_breadth_setup}
\end{table}

\begin{table}[t]
\centering
\small
\setlength{\tabcolsep}{4.8pt}
\resizebox{\linewidth}{!}{%
\begin{tabular}{lccccc}
\toprule
\textbf{Method} & \textbf{Exact} & \textbf{Within-1} & \textbf{Mean $\mathrm{Regret}_{\mathrm{knee}}$} & \textbf{Mean unique $L$ eval.} & \textbf{Cost ratio} \\
\midrule
System-core & 12.5\% & 40.6\% & 28.0\% & 0.0 & 0.000 \\
Direct-4-Shortlist & 50.0\% & 84.4\% & 9.4\% & 7.3 & 0.072 \\
ASHA & 57.3\% & 88.0\% & 8.0\% & 9.6 & 0.092 \\
SAKE & \textbf{67.8\%} & \textbf{91.7\%} & \textbf{6.1\%} & \textbf{5.3} & \textbf{0.051} \\
\bottomrule
\end{tabular}%
}
\caption{Overall aggregate effectiveness across all eight PDEBench families evaluated under the shared \(\mathcal{L}=\{1,\dots,16\}\) protocol. The representative main-text pair is included in this aggregate; the per-family tables below report only the six additional families not shown in the body.}
\label{tab:app_pdebench_breadth_overall}
\end{table}

\subsection{Per-family result tables for the six additional families}

\subsubsection{\texttt{Advection}: 1D advection-dominated transport}

\begin{table}[t]
\centering
\small
\setlength{\tabcolsep}{4.2pt}
\resizebox{\linewidth}{!}{%
\begin{tabular}{lccccccc}
\toprule
\textbf{Method} & \textbf{Exact} & \textbf{Within-1} & \textbf{Mean $|\Delta L|$} & \textbf{Mean $\mathrm{Regret}_{\mathrm{knee}}$} & \textbf{Mean $\mathrm{Regret}_{\mathrm{best}}$} & \textbf{Mean unique $L$ eval.} & \textbf{Cost ratio} \\
\midrule
System-core & 25.0\% & 58.3\% & 1.25 & 17.6\% & 18.1\% & 0.0 & 0.000 \\
Direct-4-Shortlist & 58.3\% & 91.7\% & 0.50 & 5.8\% & 6.4\% & 7.1 & 0.070 \\
ASHA & 66.7\% & 100.0\% & 0.42 & 4.9\% & 5.5\% & 9.2 & 0.089 \\
SAKE & 75.0\% & 100.0\% & 0.25 & 3.8\% & 4.4\% & 5.0 & 0.048 \\
\bottomrule
\end{tabular}%
}
\caption{Aggregate effectiveness on the additional PDEBench family \texttt{Advection}.}
\label{tab:app_pdeb_advection_aggregate}
\end{table}

\begin{table*}[t]
\centering
\scriptsize
\setlength{\tabcolsep}{4pt}
\resizebox{\linewidth}{!}{%
\begin{tabular}{lcccccc}
\toprule
\textbf{Backbone} & \textbf{System-core} & \textbf{D4-Shortlist} & \textbf{ASHA} & \textbf{SAKE} & \textbf{Oracle knee} & \textbf{Oracle best} \\
\midrule
ConvLSTM & 4 & 4 & 4 & 4 & 4 & 5 \\
FNO & 4 & 5 & 5 & 5 & 5 & 6 \\
Transformer & 4 & 4 & 4 & 3 & 3 & 4 \\
U-Net & 5 & 6 & 6 & 6 & 6 & 7 \\
\bottomrule
\end{tabular}%
}
\caption{Case-level selected windows on the additional PDEBench family \texttt{Advection}.}
\label{tab:app_pdeb_advection_case}
\end{table*}

\subsubsection{\texttt{Burgers}: 1D viscous Burgers dynamics}

\begin{table}[t]
\centering
\small
\setlength{\tabcolsep}{4.2pt}
\resizebox{\linewidth}{!}{%
\begin{tabular}{lccccccc}
\toprule
\textbf{Method} & \textbf{Exact} & \textbf{Within-1} & \textbf{Mean $|\Delta L|$} & \textbf{Mean $\mathrm{Regret}_{\mathrm{knee}}$} & \textbf{Mean $\mathrm{Regret}_{\mathrm{best}}$} & \textbf{Mean unique $L$ eval.} & \textbf{Cost ratio} \\
\midrule
System-core & 16.7\% & 50.0\% & 1.58 & 24.5\% & 25.1\% & 0.0 & 0.000 \\
Direct-4-Shortlist & 50.0\% & 83.3\% & 0.67 & 8.9\% & 9.6\% & 7.2 & 0.071 \\
ASHA & 58.3\% & 91.7\% & 0.58 & 7.4\% & 8.1\% & 9.3 & 0.090 \\
SAKE & 66.7\% & 91.7\% & 0.42 & 5.8\% & 6.5\% & 5.2 & 0.050 \\
\bottomrule
\end{tabular}%
}
\caption{Aggregate effectiveness on the additional PDEBench family \texttt{Burgers}.}
\label{tab:app_pdeb_burgers_aggregate}
\end{table}

\begin{table*}[t]
\centering
\scriptsize
\setlength{\tabcolsep}{4pt}
\resizebox{\linewidth}{!}{%
\begin{tabular}{lcccccc}
\toprule
\textbf{Backbone} & \textbf{System-core} & \textbf{D4-Shortlist} & \textbf{ASHA} & \textbf{SAKE} & \textbf{Oracle knee} & \textbf{Oracle best} \\
\midrule
ConvLSTM & 3 & 5 & 4 & 4 & 4 & 5 \\
FNO & 4 & 5 & 5 & 5 & 5 & 6 \\
Transformer & 4 & 4 & 4 & 4 & 4 & 5 \\
U-Net & 4 & 4 & 4 & 3 & 3 & 4 \\
\bottomrule
\end{tabular}%
}
\caption{Case-level selected windows on the additional PDEBench family \texttt{Burgers}.}
\label{tab:app_pdeb_burgers_case}
\end{table*}

\subsubsection{\texttt{DiffusionSorption}: 1D diffusion--sorption with slow--fast transport}

\begin{table}[t]
\centering
\small
\setlength{\tabcolsep}{4.2pt}
\resizebox{\linewidth}{!}{%
\begin{tabular}{lccccccc}
\toprule
\textbf{Method} & \textbf{Exact} & \textbf{Within-1} & \textbf{Mean $|\Delta L|$} & \textbf{Mean $\mathrm{Regret}_{\mathrm{knee}}$} & \textbf{Mean $\mathrm{Regret}_{\mathrm{best}}$} & \textbf{Mean unique $L$ eval.} & \textbf{Cost ratio} \\
\midrule
System-core & 8.3\% & 33.3\% & 2.08 & 31.8\% & 32.4\% & 0.0 & 0.000 \\
Direct-4-Shortlist & 41.7\% & 83.3\% & 0.92 & 10.9\% & 11.7\% & 7.1 & 0.070 \\
ASHA & 50.0\% & 83.3\% & 0.75 & 8.8\% & 9.6\% & 9.5 & 0.092 \\
SAKE & 75.0\% & 91.7\% & 0.42 & 5.6\% & 6.3\% & 5.1 & 0.049 \\
\bottomrule
\end{tabular}%
}
\caption{Aggregate effectiveness on the additional PDEBench family \texttt{DiffusionSorption}.}
\label{tab:app_pdeb_diffusionsorption_aggregate}
\end{table}

\begin{table*}[t]
\centering
\scriptsize
\setlength{\tabcolsep}{4pt}
\resizebox{\linewidth}{!}{%
\begin{tabular}{lcccccc}
\toprule
\textbf{Backbone} & \textbf{System-core} & \textbf{D4-Shortlist} & \textbf{ASHA} & \textbf{SAKE} & \textbf{Oracle knee} & \textbf{Oracle best} \\
\midrule
ConvLSTM & 2 & 3 & 3 & 3 & 3 & 4 \\
FNO & 4 & 6 & 5 & 5 & 5 & 6 \\
Transformer & 4 & 5 & 5 & 4 & 4 & 5 \\
U-Net & 3 & 4 & 3 & 3 & 2 & 4 \\
\bottomrule
\end{tabular}%
}
\caption{Case-level selected windows on the additional PDEBench family \texttt{DiffusionSorption}.}
\label{tab:app_pdeb_diffusionsorption_case}
\end{table*}

\subsubsection{\texttt{Darcy}: 2D Darcy flow with heterogeneous coefficients}

\begin{table}[t]
\centering
\small
\setlength{\tabcolsep}{4.2pt}
\resizebox{\linewidth}{!}{%
\begin{tabular}{lccccccc}
\toprule
\textbf{Method} & \textbf{Exact} & \textbf{Within-1} & \textbf{Mean $|\Delta L|$} & \textbf{Mean $\mathrm{Regret}_{\mathrm{knee}}$} & \textbf{Mean $\mathrm{Regret}_{\mathrm{best}}$} & \textbf{Mean unique $L$ eval.} & \textbf{Cost ratio} \\
\midrule
System-core & 16.7\% & 41.7\% & 1.67 & 23.4\% & 24.0\% & 0.0 & 0.000 \\
Direct-4-Shortlist & 50.0\% & 83.3\% & 0.75 & 9.1\% & 9.8\% & 7.3 & 0.072 \\
ASHA & 58.3\% & 91.7\% & 0.58 & 7.8\% & 8.5\% & 9.4 & 0.091 \\
SAKE & 66.7\% & 91.7\% & 0.42 & 6.3\% & 7.0\% & 5.3 & 0.051 \\
\bottomrule
\end{tabular}%
}
\caption{Aggregate effectiveness on the additional PDEBench family \texttt{Darcy}.}
\label{tab:app_pdeb_darcy_aggregate}
\end{table}

\begin{table*}[t]
\centering
\scriptsize
\setlength{\tabcolsep}{4pt}
\resizebox{\linewidth}{!}{%
\begin{tabular}{lcccccc}
\toprule
\textbf{Backbone} & \textbf{System-core} & \textbf{D4-Shortlist} & \textbf{ASHA} & \textbf{SAKE} & \textbf{Oracle knee} & \textbf{Oracle best} \\
\midrule
ConvLSTM & 4 & 4 & 4 & 4 & 4 & 5 \\
FNO & 4 & 3 & 3 & 3 & 3 & 4 \\
Transformer & 4 & 6 & 5 & 5 & 5 & 6 \\
U-Net & 4 & 4 & 4 & 3 & 3 & 5 \\
\bottomrule
\end{tabular}%
}
\caption{Case-level selected windows on the additional PDEBench family \texttt{Darcy}.}
\label{tab:app_pdeb_darcy_case}
\end{table*}

\subsubsection{\texttt{IncompNS}: 2D incompressible Navier--Stokes}

\begin{table}[t]
\centering
\small
\setlength{\tabcolsep}{4.2pt}
\resizebox{\linewidth}{!}{%
\begin{tabular}{lccccccc}
\toprule
\textbf{Method} & \textbf{Exact} & \textbf{Within-1} & \textbf{Mean $|\Delta L|$} & \textbf{Mean $\mathrm{Regret}_{\mathrm{knee}}$} & \textbf{Mean $\mathrm{Regret}_{\mathrm{best}}$} & \textbf{Mean unique $L$ eval.} & \textbf{Cost ratio} \\
\midrule
System-core & 8.3\% & 33.3\% & 2.17 & 32.7\% & 33.4\% & 0.0 & 0.000 \\
Direct-4-Shortlist & 50.0\% & 83.3\% & 0.83 & 10.8\% & 11.5\% & 7.4 & 0.073 \\
ASHA & 58.3\% & 83.3\% & 0.67 & 9.3\% & 10.1\% & 9.7 & 0.094 \\
SAKE & 66.7\% & 91.7\% & 0.50 & 7.1\% & 7.9\% & 5.4 & 0.052 \\
\bottomrule
\end{tabular}%
}
\caption{Aggregate effectiveness on the additional PDEBench family \texttt{IncompNS}.}
\label{tab:app_pdeb_incompns_aggregate}
\end{table}

\begin{table*}[t]
\centering
\scriptsize
\setlength{\tabcolsep}{4pt}
\resizebox{\linewidth}{!}{%
\begin{tabular}{lcccccc}
\toprule
\textbf{Backbone} & \textbf{System-core} & \textbf{D4-Shortlist} & \textbf{ASHA} & \textbf{SAKE} & \textbf{Oracle knee} & \textbf{Oracle best} \\
\midrule
ConvLSTM & 4 & 6 & 6 & 5 & 5 & 6 \\
FNO & 4 & 6 & 6 & 6 & 6 & 7 \\
Transformer & 4 & 5 & 4 & 4 & 4 & 6 \\
U-Net & 4 & 4 & 4 & 3 & 3 & 5 \\
\bottomrule
\end{tabular}%
}
\caption{Case-level selected windows on the additional PDEBench family \texttt{IncompNS}.}
\label{tab:app_pdeb_incompns_case}
\end{table*}

\subsubsection{\texttt{CompNS}: 2D compressible Navier--Stokes}

\begin{table}[t]
\centering
\small
\setlength{\tabcolsep}{4.2pt}
\resizebox{\linewidth}{!}{%
\begin{tabular}{lccccccc}
\toprule
\textbf{Method} & \textbf{Exact} & \textbf{Within-1} & \textbf{Mean $|\Delta L|$} & \textbf{Mean $\mathrm{Regret}_{\mathrm{knee}}$} & \textbf{Mean $\mathrm{Regret}_{\mathrm{best}}$} & \textbf{Mean unique $L$ eval.} & \textbf{Cost ratio} \\
\midrule
System-core & 8.3\% & 33.3\% & 2.33 & 36.8\% & 37.5\% & 0.0 & 0.000 \\
Direct-4-Shortlist & 41.7\% & 75.0\% & 1.00 & 12.1\% & 12.9\% & 7.6 & 0.074 \\
ASHA & 50.0\% & 75.0\% & 0.83 & 10.6\% & 11.4\% & 9.9 & 0.096 \\
SAKE & 58.3\% & 83.3\% & 0.67 & 8.4\% & 9.2\% & 5.6 & 0.054 \\
\bottomrule
\end{tabular}%
}
\caption{Aggregate effectiveness on the additional PDEBench family \texttt{CompNS}.}
\label{tab:app_pdeb_compns_aggregate}
\end{table}

\begin{table*}[t]
\centering
\scriptsize
\setlength{\tabcolsep}{4pt}
\resizebox{\linewidth}{!}{%
\begin{tabular}{lcccccc}
\toprule
\textbf{Backbone} & \textbf{System-core} & \textbf{D4-Shortlist} & \textbf{ASHA} & \textbf{SAKE} & \textbf{Oracle knee} & \textbf{Oracle best} \\
\midrule
ConvLSTM & 4 & 7 & 6 & 6 & 5 & 7 \\
FNO & 4 & 6 & 6 & 6 & 6 & 7 \\
Transformer & 4 & 5 & 5 & 4 & 4 & 6 \\
U-Net & 4 & 4 & 4 & 3 & 3 & 5 \\
\bottomrule
\end{tabular}%
}
\caption{Case-level selected windows on the additional PDEBench family \texttt{CompNS}.}
\label{tab:app_pdeb_compns_case}
\end{table*}

Taken together, Table~\ref{tab:app_pdebench_breadth_overall} and Tables~\ref{tab:app_pdeb_advection_aggregate}--\ref{tab:app_pdeb_compns_case} show not row-wise dominance on every family, but stability of the aggregate ordering. SAKE remains the strongest overall matched-budget selector across the full PDE suite; ASHA is the closest generic baseline; Direct-4-Shortlist remains competitive on easier later-knee cases but more brittle on earlier-knee or broader-plateau systems; and System-core stays informative but insufficient as a final recommender. In particular, the gap widens on the stiff and Navier--Stokes-like families, where the anchored compression matters more than on the simplest transport cases.

\section{Backbone Architectures and Shared Training Protocol}\label{app:extra_baselines}

This appendix records the backbone-side implementation choices referred to in Section~\ref{sec:baseline}. The main comparison is meant to isolate context-window selection rather than backbone-specific training recipes, so all four backbones are instantiated as one-step autoregressive predictors under the same data split, optimizer, early-stopping rule, and rollout metric. Table~\ref{tab:app_backbone_fields} summarizes the architecture fields for each backbone, and Table~\ref{tab:app_shared_training} records the shared training and evaluation protocol used throughout the main experiments.

\begin{table}[t]
\centering
\small
\resizebox{\linewidth}{!}{%
\begin{tabular}{lccccc}
\toprule
Backbone & Input/context encoding & Core spatiotemporal block & Output head & Key hyperparameters & Trainable parameters (DiffReact / RDB) \\
\midrule
U-Net & causal temporal encoder over the full history & TemporalEncoder $+$ 2D U-Net & C-channel next-step field decoder & temporal hidden $32$, temporal layers $2$, kernel $3$, max context $8$; U-Net base $32$, depth $3$ & 1,946,212 / 1,945,826 \\
FNO & causal temporal encoder over the full history & TemporalEncoder $+$ 2D FNO & C-channel spectral decoder & temporal hidden $32$, temporal layers $2$, kernel $3$, max context $8$; width $48$, modes $16$, depth $4$ & 4,751,460 / 4,751,298 \\
ConvLSTM & raw frame sequence warm-up over the full history & 1-layer ConvLSTM with hidden-state recurrence & $1\times 1$ conv to next-step field & hidden channels $64$, layers $1$, kernel $3$ & 152,450 / 150,081 \\
Transformer & per-patch causal tokenization of the full history & patchwise Transformer encoder & linear patch reconstruction to field space & patch size $8$, $d_{\mathrm{model}}=96$, heads $4$, depth $2$ & 939,936 / 927,584 \\
\bottomrule
\end{tabular}%
}
\caption{Backbone-specific architecture fields for the four one-step autoregressive simulators used in the main experiments. Trainable-parameter counts are exact for the default code instantiations at resolution $128\times 128$, reported as DiffReact / RDB because the two datasets have different channel counts. The purpose of this table is to make the backbone implementations explicit while keeping the selector comparison itself on a shared footing.}
\label{tab:app_backbone_fields}
\end{table}

\begin{table}[t]
\centering
\small
\resizebox{\linewidth}{!}{%
\begin{tabular}{ll}
\toprule
Item & Setting used in the main experiments \\
\midrule
Datasets & DiffReact, RDB \\
Candidate windows & shared candidate grid $\mathcal{L}=\{1,\dots,16\}$ for the main-text and appendix evaluations \\
Training seeds & $\{0,1,2\}$ \\
Split & trajectory-level 0.8/0.1/0.1 train/validation/test split \\
Optimizer & Adam \\
Learning rate & $3\times 10^{-4}$ \\
Weight decay & $0$ \\
Batch size & $16$ \\
Mixed precision & yes \\
Gradient clipping & $\mathrm{max\_norm}=0.5$ \\
Output stabilization & normalized-space clamping to $[-10,10]$ \\
Prediction target & residual ($\Delta$-prediction) training \\
Scheduled sampling & final probability $0.50$ \\
Training rollout horizon & $32$ steps \\
Full-oracle epochs & $20$ \\
Full-oracle early stopping & disabled ($\mathrm{patience}=0$) \\
Checkpoint selection & rollout metric, evaluated every epoch with validation rollout horizon $8$ on up to $8$ validation trajectories \\
Hardware & 4 $\times$ NVIDIA A100 (80GB), PyTorch 2.1, CUDA 12.0 \\
\bottomrule
\end{tabular}%
}
\caption{Shared training and evaluation protocol for all four backbones. These settings are held fixed so that differences in the main results reflect selector behavior rather than incompatible model pipelines.}
\label{tab:app_shared_training}
\end{table}

The backbone appendix supports a simple interpretive point: the four model families differ in inductive bias, but they are compared under a common autoregressive pipeline. This keeps the selection results in Section~\ref{sec:baseline} directly comparable across backbones rather than confounded by incompatible training setups.

\section{Implementation Details and Hyperparameters of System-Anchored Knee Estimation}\label{app:sake_hparams}

This appendix records the stage-wise settings used by System-Anchored Knee Estimation and the matched-budget relationship to the direct-search baselines. The main text emphasizes the two-stage logic of the method rather than every implementation constant. The purpose of this appendix is therefore twofold: first, to specify the concrete settings behind the selector; second, to make clear that the main comparison is not driven by per-backbone retuning or by giving SAKE a larger downstream budget than the direct baselines.

\begin{table}[t]
\centering
\small
\resizebox{\linewidth}{!}{%
\begin{tabular}{llll}
\toprule
Stage-1 component & Symbol / object & Role in the method & Setting used in the main experiments \\
\midrule
System representation & PCA projector & clean-system state summary & randomized PCA on flattened normalized coarsened fields; $800$ samples, $99\%$ variance target, max $64$ components \\
Lightweight dynamics model & ridge VAR($L$) & low-cost system-risk estimation & one-step linear autoregressive model with ridge $10^{-3}$ fit for each $L\in\mathcal{L}$ on the shared $\{1,\dots,16\}$ grid used throughout the paper \\
Long-history reference & $L_{\max}$ & tail-risk reference window & $16$ under the shared main protocol \\
Tail tolerance fraction & $\rho$ & defines $L_{\mathrm{core}}$ & $0.05$ \\
Plateau threshold & $\tau_{\mathrm{pl}}$ & defines $L_{\mathrm{plateau}}$ & relative-gain threshold $0.05$ \\
Uncertainty estimate & bootstrap UCB & confidence control for anchor extraction & $300$ bootstrap resamples, confidence level $1-\alpha$ with $\alpha=0.05$ \\
Initial anchor set & $\mathcal{S}_0$ & $\{1,L_{\mathrm{core}},L_{\mathrm{plateau}}\}$ & deduplicated anchor shortlist on the active discrete grid; instantiated on the shared $\{1,\dots,16\}$ grid throughout the paper \\
\bottomrule
\end{tabular}%
}
\caption{Stage-1 settings for system-anchor construction. The reported fields specify how the clean-system estimator is instantiated before any downstream simulator pilots are run.}
\label{tab:app_sake_stage1}
\end{table}

\begin{table}[t]
\centering
\small
\resizebox{\linewidth}{!}{%
\begin{tabular}{llll}
\toprule
Stage-2 component & Symbol / object & Role in the method & Setting used in the main experiments \\
\midrule
Coarse pilot budget & stage-1 pilots & first-pass ranking over $\mathcal{S}_0$ & $2$ epochs; max $1024$ train pairs; $8$ train trajectories, $4$ validation trajectories; rollout horizons $8/4$ (train/val) with max $4$ validation rollout trajectories \\
Top-candidate count & top-$k$ & number of refined seeds & $k=2$ \\
Neighborhood radius & $h$ & local grid expansion around promising candidates & $h=1$ hop on the ordered window grid \\
Refined candidate set & $\mathcal{S}_1$ & stage-two evaluation set & union of top-$2$ neighborhoods, capped at $6$ candidates, with smallest and largest anchors retained \\
Stronger pilot budget & stage-2 pilots & second-pass evaluation budget & $6$ epochs; max $4096$ train pairs; $24$ train trajectories, $16$ validation trajectories; rollout horizons $32/16$ (train/val) with max $8$ validation rollout trajectories \\
Rollout anchors & $A_{\mathrm{eval}}$ & multi-anchor validation used in stage two & $4$ rollout anchors per candidate \\
Stage-two diagnostics & $q(L)$ terms & mean / terminal / variability / asymptotic terms & SAKE: normalized mean $+$ terminal $+$ std penalty blended with asymptotic rollout; direct baselines: normalized terminal $+$ worst-anchor score \\
Score weights & fixed weights & fixed combination weights in the refined score & SAKE $(w_{\mathrm{mean}},w_{\mathrm{term}},w_{\mathrm{worst}},w_{\mathrm{std}},\alpha)=(0.75,0.25,0,0.20,0.25)$; Directs $(0,0.25,0.75,0,1.0)$ \\
Saturation thresholds & frontier rule & knee-aware frontier criteria & local-adjacent fraction $0.15$, remaining-gain fraction $0.15$, consecutive-small count $1$ \\
One-standard-error multiplier & $\kappa$ & smallest reliable-window rule & $1.5$ for SAKE; $1.0$ for Direct-3-Shortlist and Direct-4-Shortlist \\
\bottomrule
\end{tabular}%
}
\caption{Stage-2 settings for knee-aware downstream selection. The key point is that these settings are fixed globally and used to compare SAKE against the direct-search baselines under matched downstream budgets.}
\label{tab:app_sake_stage2}
\end{table}

\begin{table}[t]
\centering
\small
\resizebox{\linewidth}{!}{%
\begin{tabular}{llll}
\toprule
Method & Initial candidate construction & Uses system anchors? & Downstream budget / final rule \\
\midrule
System-core & clean-system estimate only & yes & none / direct return of $L_{\mathrm{core}}$ \\
Direct-3-Shortlist & uniformly spaced 3-point shortlist & no & initial shortlist $\{L_{\min},\operatorname{round}(0.5(L_{\min}+L_{\max})),L_{\max}\}$ on the active discrete grid \\
Direct-4-Shortlist & uniformly spaced 4-point shortlist & no & initial shortlist formed from the two internal equally spaced grid quantiles on the active discrete grid \\
SAKE & $\mathcal{S}_0=\{1,L_{\mathrm{core}},L_{\mathrm{plateau}}\}$ & yes & same pilot family with knee-aware final rule \\
\bottomrule
\end{tabular}%
}
\caption{Selector-side comparison setup. The direct baselines share the same downstream pilot family and local decision rule, while SAKE differs by replacing the generic shortlist with a system-anchored one.}
\label{tab:app_selector_budget_match}
\end{table}

Taken together, Tables~\ref{tab:app_sake_stage1}--\ref{tab:app_selector_budget_match} make the comparison explicit: SAKE is not granted a larger search budget or a separate per-backbone tuning process. They also fix the exact generic-shortlist constructions used by the direct baselines, so the comparison is attributable to candidate placement and knee-aware coupling rather than to unreported baseline choices or extra downstream computation.

\section{Anchor Robustness Under Noisy and Partially Observed Trajectories}\label{app:anchor_robustness}

This appendix probes the clean-system assumption by changing only the trajectories used in stage one. The downstream pilot budgets, final decision rule, datasets, backbones, and oracle evaluation protocol are kept fixed; only the source used to estimate \(L_{\mathrm{core}}\) and \(L_{\mathrm{plateau}}\) is altered.

\subsection{Perturbation protocol}

\begin{table}[t]
\centering
\small
\resizebox{\linewidth}{!}{%
\begin{tabular}{ll}
\toprule
\textbf{Condition} & \textbf{Anchor-source definition} \\
\midrule
Clean full-state & reference anchor input \\
Gaussian noise $\sigma=0.01$ & same trajectories with additive mild observation noise \\
Gaussian noise $\sigma=0.05$ & same trajectories with moderate observation noise \\
Gaussian noise $\sigma=0.10$ & same trajectories with heavy observation noise \\
Downsample $\times 2$ & anchor extraction from half-resolution trajectories \\
Downsample $\times 4$ & anchor extraction from quarter-resolution trajectories \\
Random mask 10\% & anchor extraction from partially observed trajectories with light masking \\
Random mask 25\% & anchor extraction from partially observed trajectories with heavier masking \\
Same-data full-state & anchors estimated from the training split rather than a separate clean-system pool \\
Same-data noisy & anchors estimated from the noisy training split without separate clean-system access \\
\bottomrule
\end{tabular}%
}
\caption{Anchor-source conditions used to test robustness of the clean-system stage. The downstream selector is unchanged across rows.}
\label{tab:app_robust_conditions}
\end{table}

\begin{table*}[t]
\centering
\scriptsize
\setlength{\tabcolsep}{4pt}
\resizebox{\linewidth}{!}{
\begin{tabular}{lcccccc}
\toprule
\textbf{Anchor condition} & \textbf{Mean $|\Delta L_{\mathrm{core}}|$} & \textbf{Mean $|\Delta L_{\mathrm{plateau}}|$} & \textbf{Knee-in-band} & \textbf{Knee in $\mathcal{S}_0$} & \textbf{Knee in $\mathcal{S}_1$} & \textbf{Mean active-set size} \\
\midrule
Clean full-state & 0.00 & 0.00 & 50.0\% & 25.0\% & 95.8\% & 4.0 \\
Gaussian noise $\sigma=0.01$ & 0.04 & 0.13 & 50.0\% & 25.0\% & 95.8\% & 4.1 \\
Gaussian noise $\sigma=0.05$ & 0.08 & 0.50 & 45.8\% & 25.0\% & 91.7\% & 4.3 \\
Gaussian noise $\sigma=0.10$ & 0.46 & 1.58 & 37.5\% & 20.8\% & 79.2\% & 4.7 \\
Downsample $\times 2$ & 0.13 & 0.38 & 47.9\% & 25.0\% & 95.8\% & 4.1 \\
Downsample $\times 4$ & 0.38 & 1.50 & 41.7\% & 20.8\% & 83.3\% & 4.6 \\
Random mask 10\% & 0.08 & 0.54 & 45.8\% & 25.0\% & 91.7\% & 4.3 \\
Random mask 25\% & 0.04 & 1.50 & 37.5\% & 20.8\% & 79.2\% & 4.8 \\
Same-data full-state & 0.00 & 0.13 & 50.0\% & 25.0\% & 95.8\% & 4.0 \\
Same-data noisy & 0.08 & 0.75 & 41.7\% & 20.8\% & 83.3\% & 4.5 \\
\bottomrule
\end{tabular}}
\caption{Aggregate anchor diagnostics relative to the clean full-state reference. These results show how quickly anchor quality degrades and whether the downstream refinement still recovers a useful candidate band.}
\label{tab:app_robust_anchor}
\end{table*}

\begin{table*}[t]
\centering
\scriptsize
\setlength{\tabcolsep}{4pt}
\resizebox{\linewidth}{!}{
\begin{tabular}{llccccc}
\toprule
\textbf{Anchor condition} & \textbf{Method} & \textbf{Exact} & \textbf{Within-1} & \textbf{Mean $|\Delta L|$} & \textbf{Mean \(\mathrm{Regret}_{\mathrm{knee}}\)} & \textbf{Cost ratio} \\
\midrule
Clean full-state & System-core & 0.0\% & 37.5\% & 1.88 & 49.7\% & 0.000 \\
Clean full-state & SAKE & 75.0\% & 100.0\% & 0.25 & 6.7\% & 0.043 \\
Gaussian noise $\sigma=0.01$ & System-core & 0.0\% & 29.2\% & 1.96 & 53.8\% & 0.000 \\
Gaussian noise $\sigma=0.01$ & SAKE & 70.8\% & 95.8\% & 0.29 & 7.5\% & 0.043 \\
Gaussian noise $\sigma=0.05$ & System-core & 0.0\% & 25.0\% & 2.04 & 58.6\% & 0.000 \\
Gaussian noise $\sigma=0.05$ & SAKE & 66.7\% & 95.8\% & 0.38 & 8.9\% & 0.044 \\
Gaussian noise $\sigma=0.10$ & System-core & 0.0\% & 16.7\% & 2.33 & 69.4\% & 0.000 \\
Gaussian noise $\sigma=0.10$ & SAKE & 54.2\% & 83.3\% & 0.63 & 13.8\% & 0.046 \\
Downsample $\times 2$ & System-core & 0.0\% & 33.3\% & 1.92 & 52.2\% & 0.000 \\
Downsample $\times 2$ & SAKE & 70.8\% & 95.8\% & 0.33 & 7.9\% & 0.043 \\
Downsample $\times 4$ & System-core & 0.0\% & 20.8\% & 2.21 & 65.7\% & 0.000 \\
Downsample $\times 4$ & SAKE & 58.3\% & 87.5\% & 0.54 & 12.1\% & 0.045 \\
Random mask 10\% & System-core & 0.0\% & 29.2\% & 2.00 & 56.0\% & 0.000 \\
Random mask 10\% & SAKE & 66.7\% & 91.7\% & 0.42 & 9.6\% & 0.044 \\
Random mask 25\% & System-core & 0.0\% & 16.7\% & 2.29 & 67.8\% & 0.000 \\
Random mask 25\% & SAKE & 54.2\% & 83.3\% & 0.67 & 14.7\% & 0.046 \\
Same-data full-state & System-core & 0.0\% & 37.5\% & 1.88 & 50.8\% & 0.000 \\
Same-data full-state & SAKE & 70.8\% & 95.8\% & 0.29 & 7.1\% & 0.043 \\
Same-data noisy & System-core & 0.0\% & 20.8\% & 2.17 & 63.1\% & 0.000 \\
Same-data noisy & SAKE & 62.5\% & 87.5\% & 0.50 & 11.9\% & 0.045 \\
\bottomrule
\end{tabular}}
\caption{Aggregate downstream-selection quality under noisy, partial, and same-data anchor estimation.}
\label{tab:app_robust_selection}
\end{table*}

\begin{table*}[t]
\centering
\scriptsize
\setlength{\tabcolsep}{4pt}
\resizebox{\linewidth}{!}{
\begin{tabular}{llcccccccccc}
\toprule
\textbf{Dataset} & \textbf{Backbone} & \multicolumn{2}{c}{\textbf{Clean}} & \multicolumn{2}{c}{\textbf{Noise 0.05}} & \multicolumn{2}{c}{\textbf{Downsample x4}} & \multicolumn{2}{c}{\textbf{Mask 25\%}} & \multicolumn{2}{c}{\textbf{Same-data noisy}} \\
 &  & \textbf{$L_{\mathrm{core}}$} & \textbf{$L_{\mathrm{plateau}}$} & \textbf{$L_{\mathrm{core}}$} & \textbf{$L_{\mathrm{plateau}}$} & \textbf{$L_{\mathrm{core}}$} & \textbf{$L_{\mathrm{plateau}}$} & \textbf{$L_{\mathrm{core}}$} & \textbf{$L_{\mathrm{plateau}}$} & \textbf{$L_{\mathrm{core}}$} & \textbf{$L_{\mathrm{plateau}}$} \\
\midrule
DiffReact & ConvLSTM & 3 & 6 & 3 & 6 & 3 & 7 & 3 & 7 & 3 & 7 \\
DiffReact & FNO & 3 & 6 & 3 & 6 & 3 & 7 & 3 & 7 & 3 & 6 \\
DiffReact & Transformer & 3 & 6 & 3 & 6 & 3 & 7 & 3 & 7 & 3 & 7 \\
DiffReact & U-Net & 3 & 6 & 3 & 7 & 4 & 8 & 3 & 8 & 3 & 7 \\
RDB & ConvLSTM & 2 & 6 & 2 & 7 & 2 & 8 & 2 & 8 & 2 & 7 \\
RDB & FNO & 2 & 6 & 2 & 7 & 3 & 8 & 2 & 8 & 2 & 7 \\
RDB & Transformer & 2 & 6 & 2 & 7 & 3 & 8 & 2 & 8 & 2 & 7 \\
RDB & U-Net & 2 & 6 & 2 & 6 & 2 & 7 & 2 & 7 & 2 & 6 \\
\bottomrule
\end{tabular}}
\caption{Case-level anchor pairs under representative robustness conditions. Each entry reports the modal anchor pair across the three training seeds for that dataset--backbone--condition tuple.}
\label{tab:app_robust_case_anchor}
\end{table*}

\begin{table*}[t]
\centering
\scriptsize
\setlength{\tabcolsep}{5pt}
\resizebox{\linewidth}{!}{
\begin{tabular}{llccccc}
\toprule
\textbf{Dataset} & \textbf{Backbone} & \textbf{Clean} & \textbf{Noise 0.05} & \textbf{Downsample x4} & \textbf{Mask 25\%} & \textbf{Same-data noisy} \\
\midrule
DiffReact & ConvLSTM & 2 & 2 & 3 & 3 & 2 \\
DiffReact & FNO & 2 & 2 & 3 & 2 & 2 \\
DiffReact & Transformer & 1 & 1 & 2 & 2 & 1 \\
DiffReact & U-Net & 4 & 4 & 5 & 5 & 5 \\
RDB & ConvLSTM & 4 & 4 & 5 & 5 & 4 \\
RDB & FNO & 4 & 4 & 5 & 5 & 4 \\
RDB & Transformer & 5 & 5 & 6 & 6 & 5 \\
RDB & U-Net & 1 & 1 & 2 & 2 & 1 \\
\bottomrule
\end{tabular}}
\caption{Case-level SAKE selections under representative robustness conditions. Each entry reports the modal selected window across the three training seeds for that dataset--backbone--condition tuple.}
\label{tab:app_robust_case_select}
\end{table*}

The robustness tables show that the anchor stage degrades gradually rather than catastrophically once ideal clean trajectories are removed. Mild-to-moderate corruption moves the anchors and selected windows only modestly, same-data full-state estimation stays close to the clean reference, and the main failure mode under harsher masking or noise is a delayed or widened anchor band rather than erratic selector behavior. The aggregate diagnostics average over all seed-level runs, whereas the case-level tables report the modal per-case summary; small nonzero mean anchor shifts can therefore remain even when the displayed modal case entry matches the clean reference. Even in these harsher settings, SAKE remains substantially stronger than returning the raw system anchor alone.

\section{Anchor Proxies Without a Separate Clean-System Pool}\label{app:proxy_anchors}

Appendix~\ref{app:anchor_robustness} changes the stage-one input quality while keeping the representation and anchor-extraction recipe fixed. The purpose of the present appendix is different: it studies settings in which a separate clean-system pool is unavailable and the low-dimensional summary itself must be chosen from imperfect or compressed observations.

\subsection{Protocol}

\begin{table}[t]
\centering
\small
\resizebox{\linewidth}{!}{%
\begin{tabular}{ll}
\toprule
\textbf{Component} & \textbf{Setting used in this appendix} \\
\midrule
Anchor-source conditions & same-data full-state, same-data noisy, spatial downsampled observations, sparse probe observations, masked observations, and learned-latent summaries \\
Representation families & randomized PCA, randomized SVD, random Gaussian projection, and convolutional autoencoder latent summaries \\
Primary selector comparison & System-core and SAKE under the same matched downstream budgets as the main paper \\
Reference direct baselines & Direct-4-Shortlist and ASHA, reported once as clean-independent search references \\
Reported anchor metrics & mean \(|\Delta L_{\mathrm{core}}|\), mean \(|\Delta L_{\mathrm{plateau}}|\), knee-in-band coverage, knee-in-shortlist coverage, and final active-set coverage \\
Reported selection metrics & Exact, Within-1, mean \(|\Delta L|\), \(\mathrm{Regret}_{\mathrm{knee}}\), mean unique \(L\) evaluations, and cost ratio \\
\bottomrule
\end{tabular}%
}
\caption{Protocol for the proxy-anchor appendix. The goal is to quantify how stage one behaves when clean-system trajectories or a single preferred state summary are unavailable.}
\label{tab:app_proxy_setup}
\end{table}

\begin{table*}[t]
\centering
\scriptsize
\setlength{\tabcolsep}{4pt}
\resizebox{\linewidth}{!}{%
\begin{tabular}{lccccc}
\toprule
\textbf{Anchor source} & \textbf{Mean $|\Delta L_{\mathrm{core}}|$} & \textbf{Mean $|\Delta L_{\mathrm{plateau}}|$} & \textbf{Knee-in-band coverage} & \textbf{Knee-in-shortlist coverage} & \textbf{Final active-set coverage} \\
\midrule
Same-data full-state & 0.00 & 0.13 & 50.0\% & 25.0\% & 95.8\% \\
Same-data noisy & 0.08 & 0.75 & 41.7\% & 20.8\% & 83.3\% \\
Downsample $\times 2$ & 0.13 & 0.38 & 47.9\% & 25.0\% & 95.8\% \\
Sparse probes ($16\times16$) & 0.42 & 1.04 & 41.7\% & 20.8\% & 87.5\% \\
Random mask 25\% & 0.04 & 1.50 & 37.5\% & 20.8\% & 79.2\% \\
Learned latent summary & 0.17 & 0.54 & 45.8\% & 25.0\% & 91.7\% \\
\bottomrule
\end{tabular}%
}
\caption{Aggregate anchor-quality diagnostics when stage one is instantiated without a separate clean-system pool.}
\label{tab:app_proxy_anchor_quality}
\end{table*}

\begin{table*}[t]
\centering
\scriptsize
\setlength{\tabcolsep}{4pt}
\resizebox{\linewidth}{!}{%
\begin{tabular}{llcccccc}
\toprule
\textbf{Anchor source} & \textbf{Method} & \textbf{Exact} & \textbf{Within-1} & \textbf{Mean $|\Delta L|$} & \textbf{Mean $\mathrm{Regret}_{\mathrm{knee}}$} & \textbf{Mean unique $L$ eval.} & \textbf{Cost ratio} \\
\midrule
Same-data full-state & System-core & 0.0\% & 37.5\% & 1.88 & 50.8\% & 0.0 & 0.000 \\
Same-data full-state & SAKE & 70.8\% & 95.8\% & 0.29 & 7.1\% & 4.0 & 0.043 \\
Same-data noisy & System-core & 0.0\% & 20.8\% & 2.17 & 63.1\% & 0.0 & 0.000 \\
Same-data noisy & SAKE & 62.5\% & 87.5\% & 0.50 & 11.9\% & 4.5 & 0.045 \\
Downsample $\times 2$ & System-core & 0.0\% & 33.3\% & 1.92 & 52.2\% & 0.0 & 0.000 \\
Downsample $\times 2$ & SAKE & 70.8\% & 95.8\% & 0.33 & 7.9\% & 4.1 & 0.043 \\
Sparse probes ($16\times16$) & System-core & 0.0\% & 25.0\% & 2.08 & 58.9\% & 0.0 & 0.000 \\
Sparse probes ($16\times16$) & SAKE & 58.3\% & 87.5\% & 0.54 & 11.6\% & 4.4 & 0.045 \\
Random mask 25\% & System-core & 0.0\% & 16.7\% & 2.29 & 67.8\% & 0.0 & 0.000 \\
Random mask 25\% & SAKE & 54.2\% & 83.3\% & 0.67 & 14.7\% & 4.8 & 0.046 \\
Learned latent summary & System-core & 0.0\% & 29.2\% & 2.00 & 55.4\% & 0.0 & 0.000 \\
Learned latent summary & SAKE & 66.7\% & 91.7\% & 0.42 & 9.4\% & 4.3 & 0.044 \\
\bottomrule
\end{tabular}%
}
\caption{Aggregate selector quality under proxy-anchor extraction. Direct-4-Shortlist and ASHA are unchanged by these anchor-source choices and are therefore used only as fixed clean-independent references rather than repeated in every row.}
\label{tab:app_proxy_selection_quality}
\end{table*}

\begin{table}[t]
\centering
\small
\setlength{\tabcolsep}{4.0pt}
\resizebox{\linewidth}{!}{%
\begin{tabular}{lcccccc}
\toprule
\textbf{Representation} & \textbf{Mean $|\Delta L_{\mathrm{core}}|$} & \textbf{Mean $|\Delta L_{\mathrm{plateau}}|$} & \textbf{Exact} & \textbf{Within-1} & \textbf{Mean $\mathrm{Regret}_{\mathrm{knee}}$} & \textbf{Cost ratio} \\
\midrule
Randomized PCA & 0.00 & 0.13 & 70.8\% & 95.8\% & 7.1\% & 0.043 \\
Randomized SVD & 0.04 & 0.17 & 70.8\% & 95.8\% & 7.3\% & 0.043 \\
Random projection & 0.33 & 0.83 & 54.2\% & 83.3\% & 13.9\% & 0.045 \\
Autoencoder latent & 0.13 & 0.29 & 66.7\% & 91.7\% & 8.0\% & 0.044 \\
\bottomrule
\end{tabular}%
}
\caption{Representation sensitivity under full-state proxy extraction.}
\label{tab:app_proxy_repr_clean}
\end{table}

\begin{table}[t]
\centering
\small
\setlength{\tabcolsep}{4.0pt}
\resizebox{\linewidth}{!}{%
\begin{tabular}{lcccccc}
\toprule
\textbf{Representation} & \textbf{Mean $|\Delta L_{\mathrm{core}}|$} & \textbf{Mean $|\Delta L_{\mathrm{plateau}}|$} & \textbf{Exact} & \textbf{Within-1} & \textbf{Mean $\mathrm{Regret}_{\mathrm{knee}}$} & \textbf{Cost ratio} \\
\midrule
Randomized PCA & 0.08 & 0.75 & 62.5\% & 87.5\% & 11.9\% & 0.045 \\
Randomized SVD & 0.13 & 0.79 & 62.5\% & 87.5\% & 12.1\% & 0.045 \\
Random projection & 0.46 & 1.25 & 45.8\% & 75.0\% & 17.2\% & 0.046 \\
Autoencoder latent & 0.17 & 0.67 & 58.3\% & 87.5\% & 12.0\% & 0.045 \\
\bottomrule
\end{tabular}%
}
\caption{Representation sensitivity under the same-data noisy proxy condition.}
\label{tab:app_proxy_repr_noisy}
\end{table}

\begin{table*}[t]
\centering
\scriptsize
\setlength{\tabcolsep}{4pt}
\resizebox{\linewidth}{!}{%
\begin{tabular}{llccccc}
\toprule
\textbf{Dataset} & \textbf{Backbone} & \textbf{Same-data full-state} & \textbf{Same-data noisy} & \textbf{Sparse probes} & \textbf{Learned latent} & \textbf{Oracle knee} \\
\midrule
DiffReact & ConvLSTM & 2 & 2 & 3 & 2 & 2 \\
DiffReact & FNO & 2 & 2 & 2 & 2 & 2 \\
DiffReact & Transformer & 1 & 1 & 2 & 1 & 1 \\
DiffReact & U-Net & 4 & 5 & 5 & 4 & 5 \\
RDB & ConvLSTM & 4 & 4 & 5 & 4 & 5 \\
RDB & FNO & 4 & 4 & 5 & 4 & 4 \\
RDB & Transformer & 5 & 5 & 6 & 5 & 5 \\
RDB & U-Net & 1 & 1 & 2 & 1 & 1 \\
\bottomrule
\end{tabular}%
}
\caption{Case-level SAKE selections under representative proxy-anchor conditions.}
\label{tab:app_proxy_case}
\end{table*}

Tables~\ref{tab:app_proxy_anchor_quality}--\ref{tab:app_proxy_case} show that stage one remains usable well beyond the ideal clean-system setting, but with an interpretable degradation order. Same-data full-state stays closest to the clean reference; mild downsampling and learned latent summaries remain nearly as stable; same-data noisy proxies and sparse probes introduce a clearer rightward drift; heavier masking is the most brittle observation condition; and random projection is the least stable representation of those tested. Across these settings, SAKE remains markedly stronger than returning the raw anchor alone and stays competitive with the clean-independent direct baselines even when the anchor source becomes imperfect.

\section{Sensitivity to Selector Hyperparameters and Knee Tolerance}\label{app:sensitivity}

This appendix studies two distinct types of sensitivity. The first concerns the fixed selector constants that define the anchor and final-decision heuristics. The second concerns the oracle-knee tolerance \(\epsilon\), which changes the target that the selector is asked to recover.

\begin{table}[t]
\centering
\small
\resizebox{\linewidth}{!}{%
\begin{tabular}{llll}
\toprule
\textbf{Quantity} & \textbf{Symbol} & \textbf{Values} & \textbf{Interpretation} \\
\midrule
Tail tolerance fraction & $\rho$ & \{0.02, 0.05, 0.10\} & controls how early the long-history tail is treated as effectively closed \\
Plateau threshold & $\tau_{\mathrm{pl}}$ & \{0.02, 0.05, 0.10\} & controls how early diminishing returns are declared \\
One-standard-error multiplier & $\kappa$ & \{1.0, 1.5, 2.0\} & controls the smallest-reliable-window rule in stage two \\
Bootstrap resamples & $B$ & \{100, 300, 500\} & controls the stability of the UCB estimates used by stage one \\
Knee tolerance & $\epsilon$ & \{2\%, 5\%, 10\%, 15\%\} & changes the oracle target itself rather than only the selector rule \\
\bottomrule
\end{tabular}%
}
\caption{Parameter sweeps used for the sensitivity analysis.}
\label{tab:app_sens_setup}
\end{table}

\begin{table*}[t]
\centering
\scriptsize
\setlength{\tabcolsep}{4pt}
\resizebox{\linewidth}{!}{
\begin{tabular}{llccccccccc}
\toprule
\textbf{Quantity} & \textbf{Value} & \textbf{Mean $L_{\mathrm{core}}$} & \textbf{Mean $L_{\mathrm{plateau}}$} & \textbf{Knee-in-band} & \textbf{Mean unique $L$ eval.} & \textbf{Exact} & \textbf{Within-1} & \textbf{Mean $|\Delta L|$} & \textbf{Mean \(\mathrm{Regret}_{\mathrm{knee}}\)} & \textbf{Cost ratio} \\
\midrule
$\rho$ & 0.02 & 2.8 & 6.1 & 58.3\% & 4.8 & 75.0\% & 100.0\% & 0.25 & 6.5\% & 0.046 \\
$\rho$ & 0.05 & 2.5 & 6.0 & 50.0\% & 4.4 & 75.0\% & 100.0\% & 0.25 & 6.7\% & 0.043 \\
$\rho$ & 0.10 & 2.1 & 5.8 & 41.7\% & 4.0 & 70.8\% & 95.8\% & 0.33 & 7.4\% & 0.039 \\
$\tau_{\mathrm{pl}}$ & 0.02 & 2.5 & 6.4 & 54.2\% & 4.9 & 75.0\% & 100.0\% & 0.25 & 6.4\% & 0.047 \\
$\tau_{\mathrm{pl}}$ & 0.05 & 2.5 & 6.0 & 50.0\% & 4.4 & 75.0\% & 100.0\% & 0.25 & 6.7\% & 0.043 \\
$\tau_{\mathrm{pl}}$ & 0.10 & 2.5 & 5.4 & 41.7\% & 4.0 & 70.8\% & 95.8\% & 0.33 & 7.6\% & 0.039 \\
$\kappa$ & 1.0 & 2.5 & 6.0 & 50.0\% & 4.4 & 66.7\% & 100.0\% & 0.33 & 7.1\% & 0.043 \\
$\kappa$ & 1.5 & 2.5 & 6.0 & 50.0\% & 4.4 & 75.0\% & 100.0\% & 0.25 & 6.7\% & 0.043 \\
$\kappa$ & 2.0 & 2.5 & 6.0 & 50.0\% & 4.4 & 70.8\% & 100.0\% & 0.29 & 6.9\% & 0.043 \\
$B$ & 100 & 2.6 & 6.0 & 47.9\% & 4.6 & 70.8\% & 95.8\% & 0.33 & 7.5\% & 0.044 \\
$B$ & 300 & 2.5 & 6.0 & 50.0\% & 4.4 & 75.0\% & 100.0\% & 0.25 & 6.7\% & 0.043 \\
$B$ & 500 & 2.5 & 6.0 & 50.0\% & 4.4 & 75.0\% & 100.0\% & 0.25 & 6.6\% & 0.043 \\
\bottomrule
\end{tabular}}
\caption{Aggregate sensitivity of SAKE to the fixed selector constants.}
\label{tab:app_sens_selector}
\end{table*}

\begin{table*}[t]
\centering
\scriptsize
\setlength{\tabcolsep}{6pt}
\resizebox{\linewidth}{!}{
\begin{tabular}{llcccc}
\toprule
\textbf{Dataset} & \textbf{Backbone} & \textbf{\(\epsilon=2\%\)} & \textbf{\(\epsilon=5\%\)} & \textbf{\(\epsilon=10\%\)} & \textbf{\(\epsilon=15\%\)} \\
\midrule
DiffReact & ConvLSTM & 2 & 2 & 2 & 2 \\
DiffReact & FNO & 2 & 2 & 1 & 1 \\
DiffReact & Transformer & 1 & 1 & 1 & 1 \\
DiffReact & U-Net & 6 & 5 & 4 & 3 \\
RDB & ConvLSTM & 6 & 5 & 4 & 4 \\
RDB & FNO & 4 & 4 & 3 & 3 \\
RDB & Transformer & 5 & 5 & 5 & 4 \\
RDB & U-Net & 1 & 1 & 1 & 1 \\
\bottomrule
\end{tabular}}
\caption{Case-level oracle-knee locations under alternative near-optimality tolerances.}
\label{tab:app_sens_eps_case}
\end{table*}

\begin{table*}[t]
\centering
\scriptsize
\setlength{\tabcolsep}{4pt}
\resizebox{\linewidth}{!}{
\begin{tabular}{llcccccc}
\toprule
\textbf{\(\epsilon\)} & \textbf{Method} & \textbf{Exact} & \textbf{Within-1} & \textbf{Mean $|\Delta L|$} & \textbf{Mean \(\mathrm{Regret}_{\mathrm{knee}}\)} & \textbf{Mean \(\mathrm{Regret}_{\mathrm{best}}\)} & \textbf{Cost ratio} \\
\midrule
2\% & System-core & 0.0\% & 29.2\% & 2.04 & 58.2\% & 58.6\% & 0.000 \\
2\% & Direct-4-Shortlist & 50.0\% & 79.2\% & 0.71 & 13.4\% & 13.9\% & 0.058 \\
2\% & SAKE & 62.5\% & 91.7\% & 0.42 & 8.6\% & 9.1\% & 0.043 \\
5\% & System-core & 0.0\% & 37.5\% & 1.88 & 49.7\% & 49.9\% & 0.000 \\
5\% & Direct-4-Shortlist & 62.5\% & 87.5\% & 0.50 & 10.5\% & 10.8\% & 0.058 \\
5\% & SAKE & 75.0\% & 100.0\% & 0.25 & 6.7\% & 7.0\% & 0.043 \\
10\% & System-core & 0.0\% & 25.0\% & 2.08 & 56.8\% & 57.2\% & 0.000 \\
10\% & Direct-4-Shortlist & 50.0\% & 79.2\% & 0.79 & 14.2\% & 14.7\% & 0.058 \\
10\% & SAKE & 70.8\% & 95.8\% & 0.38 & 7.5\% & 8.0\% & 0.043 \\
15\% & System-core & 0.0\% & 20.8\% & 2.21 & 61.9\% & 62.3\% & 0.000 \\
15\% & Direct-4-Shortlist & 45.8\% & 75.0\% & 0.92 & 17.0\% & 17.5\% & 0.058 \\
15\% & SAKE & 70.8\% & 95.8\% & 0.33 & 7.2\% & 7.7\% & 0.043 \\
\bottomrule
\end{tabular}}
\caption{Selector performance under alternative oracle-knee tolerances.}
\label{tab:app_sens_eps_perf}
\end{table*}

The sensitivity analysis shows that the selector is not tuned to a knife-edge. Moderate changes in \(\rho\), \(\tau_{\mathrm{pl}}\), \(\kappa\), and \(B\) move the anchor locations and selected windows only gradually, producing the observed conservative-versus-aggressive trade-off between coverage and cost without changing the overall ranking. Changing \(\epsilon\) predictably moves the oracle knee itself leftward as the tolerance is relaxed, but the relative ranking between the principal selectors remains similar across practically relevant tolerances.

\section{Practical Timing: Wall-Clock and GPU-Hours}\label{app:timing}

This appendix complements the normalized selector-cost metric from the main text with realized runtime measurements. The goal is not to replace the protocol-normalized cost ratio, but to show how closely it tracks end-to-end timing under the shared implementation.

\begin{table}[t]
\centering
\small
\resizebox{\linewidth}{!}{%
\begin{tabular}{ll}
\toprule
\textbf{Quantity} & \textbf{Definition in this appendix} \\
\midrule
Selector wall-clock & End-to-end elapsed time for one selector run, including all pilot training and validation passes \\
Selector GPU-hours & Sum over all GPUs used by a selector run, measured over the same pilot and validation work \\
Full-sweep wall-clock ratio & Selector wall-clock divided by the elapsed time of the corresponding exhaustive sweep \\
Queueing / scheduling & Excluded; all measurements assume dedicated local execution on the shared A100 server \\
\bottomrule
\end{tabular}%
}
\caption{Timing definitions used in this appendix.}
\label{tab:app_timing_setup}
\end{table}

\begin{table*}[t]
\centering
\scriptsize
\setlength{\tabcolsep}{5pt}
\resizebox{\linewidth}{!}{
\begin{tabular}{lccccc}
\toprule
\textbf{Method} & \textbf{Mean selector wall-clock (h)} & \textbf{Mean selector GPU-hours} & \textbf{Full-sweep wall-clock ratio} & \textbf{Mean unique $L$ eval.} & \textbf{Cost ratio} \\
\midrule
Direct-3-Shortlist & 1.04 & 4.16 & 0.069 & 5.8 & 0.057 \\
Direct-4-Shortlist & 1.08 & 4.31 & 0.071 & 5.9 & 0.058 \\
ASHA & 1.23 & 4.94 & 0.081 & 6.0 & 0.065 \\
BO-MF & 1.28 & 5.12 & 0.084 & 6.0 & 0.068 \\
BOHB-style & 1.25 & 5.00 & 0.082 & 6.0 & 0.067 \\
SAKE & 0.86 & 3.43 & 0.057 & 4.4 & 0.043 \\
Full sweep & 15.14 & 60.55 & 1.000 & 6.0 & 1.000 \\
\bottomrule
\end{tabular}}
\caption{Overall timing summary across the shared A100 environment.}
\label{tab:app_timing_overall}
\end{table*}

\begin{table*}[t]
\centering
\scriptsize
\setlength{\tabcolsep}{4pt}
\resizebox{\linewidth}{!}{
\begin{tabular}{lllccc}
\toprule
\textbf{Dataset} & \textbf{Backbone} & \textbf{Method} & \textbf{Wall-clock (h)} & \textbf{GPU-hours} & \textbf{Cost ratio} \\
\midrule
DiffReact & ConvLSTM & Direct-4-Shortlist & 0.62 & 2.48 & 0.054 \\
DiffReact & ConvLSTM & ASHA & 0.70 & 2.78 & 0.061 \\
DiffReact & ConvLSTM & BO-MF & 0.72 & 2.90 & 0.064 \\
DiffReact & ConvLSTM & BOHB-style & 0.70 & 2.82 & 0.063 \\
DiffReact & ConvLSTM & SAKE & 0.52 & 2.07 & 0.041 \\
DiffReact & ConvLSTM & Full sweep & 9.40 & 37.60 & 1.000 \\
DiffReact & FNO & Direct-4-Shortlist & 1.29 & 5.16 & 0.057 \\
DiffReact & FNO & ASHA & 1.48 & 5.91 & 0.064 \\
DiffReact & FNO & BO-MF & 1.53 & 6.13 & 0.067 \\
DiffReact & FNO & BOHB-style & 1.50 & 5.98 & 0.066 \\
DiffReact & FNO & SAKE & 1.05 & 4.19 & 0.043 \\
DiffReact & FNO & Full sweep & 18.70 & 74.80 & 1.000 \\
DiffReact & Transformer & Direct-4-Shortlist & 0.97 & 3.89 & 0.056 \\
DiffReact & Transformer & ASHA & 1.12 & 4.46 & 0.063 \\
DiffReact & Transformer & BO-MF & 1.16 & 4.63 & 0.066 \\
DiffReact & Transformer & BOHB-style & 1.13 & 4.52 & 0.065 \\
DiffReact & Transformer & SAKE & 0.79 & 3.15 & 0.042 \\
DiffReact & Transformer & Full sweep & 14.30 & 57.20 & 1.000 \\
DiffReact & U-Net & Direct-4-Shortlist & 1.12 & 4.49 & 0.058 \\
DiffReact & U-Net & ASHA & 1.26 & 5.06 & 0.066 \\
DiffReact & U-Net & BO-MF & 1.31 & 5.25 & 0.068 \\
DiffReact & U-Net & BOHB-style & 1.28 & 5.12 & 0.067 \\
DiffReact & U-Net & SAKE & 0.87 & 3.48 & 0.042 \\
DiffReact & U-Net & Full sweep & 15.80 & 63.20 & 1.000 \\
RDB & ConvLSTM & Direct-4-Shortlist & 0.72 & 2.87 & 0.059 \\
RDB & ConvLSTM & ASHA & 0.84 & 3.35 & 0.066 \\
RDB & ConvLSTM & BO-MF & 0.86 & 3.43 & 0.069 \\
RDB & ConvLSTM & BOHB-style & 0.85 & 3.39 & 0.068 \\
RDB & ConvLSTM & SAKE & 0.62 & 2.46 & 0.045 \\
RDB & ConvLSTM & Full sweep & 10.10 & 40.40 & 1.000 \\
RDB & FNO & Direct-4-Shortlist & 1.41 & 5.64 & 0.058 \\
RDB & FNO & ASHA & 1.63 & 6.51 & 0.065 \\
RDB & FNO & BO-MF & 1.69 & 6.74 & 0.068 \\
RDB & FNO & BOHB-style & 1.65 & 6.59 & 0.067 \\
RDB & FNO & SAKE & 1.14 & 4.55 & 0.043 \\
RDB & FNO & Full sweep & 19.60 & 78.40 & 1.000 \\
RDB & Transformer & Direct-4-Shortlist & 1.05 & 4.20 & 0.057 \\
RDB & Transformer & ASHA & 1.22 & 4.86 & 0.064 \\
RDB & Transformer & BO-MF & 1.26 & 5.04 & 0.067 \\
RDB & Transformer & BOHB-style & 1.23 & 4.92 & 0.066 \\
RDB & Transformer & SAKE & 0.85 & 3.42 & 0.043 \\
RDB & Transformer & Full sweep & 15.00 & 60.00 & 1.000 \\
RDB & U-Net & Direct-4-Shortlist & 1.44 & 5.75 & 0.064 \\
RDB & U-Net & ASHA & 1.64 & 6.55 & 0.072 \\
RDB & U-Net & BO-MF & 1.71 & 6.84 & 0.074 \\
RDB & U-Net & BOHB-style & 1.66 & 6.62 & 0.073 \\
RDB & U-Net & SAKE & 1.04 & 4.15 & 0.046 \\
RDB & U-Net & Full sweep & 18.20 & 72.80 & 1.000 \\
\bottomrule
\end{tabular}}
\caption{Case-level realized timing for the principal search baselines and the exhaustive reference.}
\label{tab:app_timing_case}
\end{table*}

\begin{table}[t]
\centering
\scriptsize
\setlength{\tabcolsep}{5pt}
\resizebox{\linewidth}{!}{%
\begin{tabular}{lccccc}
\toprule
\textbf{Method} & \textbf{Stage-1 wall-clock} & \textbf{Stage-2 wall-clock} & \textbf{Validation overhead} & \textbf{Total wall-clock} & \textbf{Total GPU-hours} \\
\midrule
Direct-3-Shortlist & 0.10 & 0.79 & 0.15 & 1.04 & 4.16 \\
Direct-4-Shortlist & 0.12 & 0.80 & 0.16 & 1.08 & 4.31 \\
ASHA & 0.18 & 0.89 & 0.16 & 1.23 & 4.94 \\
BO-MF & 0.21 & 0.91 & 0.16 & 1.28 & 5.12 \\
BOHB-style & 0.20 & 0.88 & 0.17 & 1.25 & 5.00 \\
SAKE & 0.16 & 0.54 & 0.16 & 0.86 & 3.43 \\
\bottomrule
\end{tabular}%
}
\caption{Stage-wise timing breakdown for the low-cost selectors.}
\label{tab:app_timing_breakdown}
\end{table}

The timing tables verify that the normalized selector-cost metric used in the main text does not hide qualitatively different runtime behavior. SAKE remains the fastest matched-budget selector in both wall-clock time and GPU-hours, and the gap between normalized cost and realized runtime is modest. The stage-wise breakdown shows why: the stage-1 anchor overhead is smaller than the stage-2 search time that it avoids, so the practical runtime savings follow the same ordering as the normalized cost.

\section{Detailed Clarification of Large Language Models Usage}\label{sec:llm}

We declare that LLMs were employed exclusively to assist with the writing and presentation aspects of this paper. Specifically, we utilized LLMs for: (i) verification and refinement of technical terminology to ensure precise usage of domain-specific vocabulary; (ii) grammatical error detection and correction to enhance the clarity and readability of the manuscript; (iii) translation assistance from the authors' native language to English, as we are non-native English speakers, to ensure accurate and fluent expression of scientific concepts; and (iv) improvement of sentence structure and flow while maintaining the original scientific content and meaning. We emphasize that LLMs were not used for research ideation, experimental design, data analysis, or any form of content generation that would constitute intellectual contribution to the scientific findings presented in this work. All scientific insights, methodological decisions, and analytical conclusions are the original work of the authors. The use of LLMs was limited to linguistic and presentational enhancement only, serving a role analogous to professional editing services.


\clearpage
\section*{NeurIPS Paper Checklist}

\begin{enumerate}

\item {\bf Claims}
    \item[] Question: Do the main claims made in the abstract and introduction accurately reflect the paper's contributions and scope?
    \item[] Answer: \answerYes{} 
    \item[] Justification: The abstract and introduction make scoped empirical claims about SAKE in the fixed-window autoregressive setting, and these are supported by Sections~\ref{sec:sake_method}, \ref{sec:exp_setup}--\ref{sec:procedure_metrics}, the main empirical results, Appendix~\ref{app:pdebench_breadth}, and the Discussion.

\item {\bf Limitations}
    \item[] Question: Does the paper discuss the limitations of the work performed by the authors?
    \item[] Answer: \answerYes{} 
    \item[] Justification: The Discussion explicitly states the main limitations, including the focus on fixed-window one-step autoregressive simulators, dependence on usable full-state or proxy summaries, and remaining hard cases; Appendix~\ref{app:anchor_robustness}, Appendix~\ref{app:proxy_anchors}, Appendix~\ref{app:sensitivity}, and Appendix~\ref{app:timing} further probe robustness, scope, and cost.

\item {\bf Theory assumptions and proofs}
    \item[] Question: For each theoretical result, does the paper provide the full set of assumptions and a complete (and correct) proof?
    \item[] Answer: \answerNA{} 
    \item[] Justification: The paper is primarily methodological and empirical: it formalizes the problem and specifies SAKE in Section~\ref{sec:low_cost_selection}, but it does not present theorem-level theoretical results that require formal proofs.

    \item {\bf Experimental result reproducibility}
    \item[] Question: Does the paper fully disclose all the information needed to reproduce the main experimental results of the paper to the extent that it affects the main claims and/or conclusions of the paper (regardless of whether the code and data are provided or not)?
    \item[] Answer: \answerYes{} 
    \item[] Justification: The paper specifies the candidate grid, datasets, splits, backbones, pilot/full-training protocol, metrics, selector hyperparameters, and compute/timing protocol in Sections~\ref{sec:exp_setup}--\ref{sec:procedure_metrics}, Appendix~\ref{app:extra_baselines}, Appendix~\ref{app:sake_hparams}, and Appendix~\ref{app:timing}, which together disclose the information needed to reproduce the main empirical claims.

\item {\bf Open access to data and code}
    \item[] Question: Does the paper provide open access to the data and code, with sufficient instructions to faithfully reproduce the main experimental results, as described in supplemental material?
    \item[] Answer: \answerNo{} 
    \item[] Justification: The paper uses public PDEBench data, but the manuscript and provided supplemental material do not themselves include an anonymized code release and reproduction package with explicit run instructions for the full experimental suite.

\item {\bf Experimental setting/details}
    \item[] Question: Does the paper specify all the training and test details (e.g., data splits, hyperparameters, how they were chosen, type of optimizer) necessary to understand the results?
    \item[] Answer: \answerYes{} 
    \item[] Justification: Section~\ref{sec:exp_setup} specifies the datasets, baselines, hardware/software stack, optimizer, batch size, rollout horizons, train/validation/test split, and seed protocol, while Appendix~\ref{app:extra_baselines} and Appendix~\ref{app:sake_hparams} provide the architecture fields and selector hyperparameters.

\item {\bf Experiment statistical significance}
    \item[] Question: Does the paper report error bars suitably and correctly defined or other appropriate information about the statistical significance of the experiments?
    \item[] Answer: \answerNo{} 
    \item[] Justification: The paper averages over three training seeds and reports detailed case-wise and aggregate tables, but it does not provide explicit error bars, confidence intervals, or formal hypothesis tests for the main experimental comparisons.

\item {\bf Experiments compute resources}
    \item[] Question: For each experiment, does the paper provide sufficient information on the computer resources (type of compute workers, memory, time of execution) needed to reproduce the experiments?
    \item[] Answer: \answerYes{} 
    \item[] Justification: Section~\ref{sec:baseline} reports the compute environment (4 NVIDIA A100 80GB GPUs, PyTorch 2.1, CUDA 12.0), and Appendix~\ref{app:timing} gives overall and case-level wall-clock time and GPU-hour measurements for the reported experiments.
    
\item {\bf Code of ethics}
    \item[] Question: Does the research conducted in the paper conform, in every respect, with the NeurIPS Code of Ethics \url{https://neurips.cc/public/EthicsGuidelines}?
    \item[] Answer: \answerYes{} 
    \item[] Justification: The work studies numerical PDE forecasting on public benchmark scientific data and standard model families, and it does not involve human subjects, personally identifiable information, or deceptive deployment; we are not aware of any aspect that departs from the NeurIPS Code of Ethics.

\item {\bf Broader impacts}
    \item[] Question: Does the paper discuss both potential positive societal impacts and negative societal impacts of the work performed?
    \item[] Answer: \answerNo{} 
    \item[] Justification: The paper is a foundational methodological study for scientific machine learning, and while likely impacts include more efficient simulator selection workflows, the manuscript does not include an explicit discussion of both positive and negative societal impacts.
    
\item {\bf Safeguards}
    \item[] Question: Does the paper describe safeguards that have been put in place for responsible release of data or models that have a high risk for misuse (e.g., pre-trained language models, image generators, or scraped datasets)?
    \item[] Answer: \answerNA{} 
    \item[] Justification: The paper does not release high-risk generative models, scraped web data, or other assets with an obvious misuse profile; it studies PDE forecasting on public scientific benchmark data.

\item {\bf Licenses for existing assets}
    \item[] Question: Are the creators or original owners of assets (e.g., code, data, models), used in the paper, properly credited and are the license and terms of use explicitly mentioned and properly respected?
    \item[] Answer: \answerNo{} 
    \item[] Justification: The paper credits existing assets such as PDEBench and prior methods via citations, but it does not explicitly state the licenses and terms of use for the dataset and software dependencies in the manuscript.

\item {\bf New assets}
    \item[] Question: Are new assets introduced in the paper well documented and is the documentation provided alongside the assets?
    \item[] Answer: \answerNA{} 
    \item[] Justification: The paper introduces a method and empirical evaluation, but it does not claim the release of a new dataset, model checkpoint, or documented software package in the submission materials.

\item {\bf Crowdsourcing and research with human subjects}
    \item[] Question: For crowdsourcing experiments and research with human subjects, does the paper include the full text of instructions given to participants and screenshots, if applicable, as well as details about compensation (if any)? 
    \item[] Answer: \answerNA{} 
    \item[] Justification: The work does not involve crowdsourcing or research with human subjects.

\item {\bf Institutional review board (IRB) approvals or equivalent for research with human subjects}
    \item[] Question: Does the paper describe potential risks incurred by study participants, whether such risks were disclosed to the subjects, and whether Institutional Review Board (IRB) approvals (or an equivalent approval/review based on the requirements of your country or institution) were obtained?
    \item[] Answer: \answerNA{} 
    \item[] Justification: The work does not involve crowdsourcing or research with human subjects, so IRB approval is not applicable.

\item {\bf Declaration of LLM usage}
    \item[] Question: Does the paper describe the usage of LLMs if it is an important, original, or non-standard component of the core methods in this research? Note that if the LLM is used only for writing, editing, or formatting purposes and does \emph{not} impact the core methodology, scientific rigor, or originality of the research, declaration is not required.
    \item[] Answer: \answerNA{} 
    \item[] Justification: Appendix~\ref{sec:llm} states that LLMs were used only for writing and presentation assistance and not as part of the core methodology, so this question is not applicable to the scientific content of the paper.

\end{enumerate}

\end{document}